\documentclass[conference]{IEEEtran}
\IEEEoverridecommandlockouts

\usepackage{cite}
\usepackage{amsmath,amssymb,amsfonts}
\usepackage{algorithmic}
\usepackage{graphicx}
\usepackage{textcomp}
\usepackage{xcolor}
\usepackage{tabularx}
\usepackage{multirow}
\usepackage{float}
\usepackage{cite}
\usepackage{multirow}
\usepackage{makecell}
\usepackage{caption}
\usepackage{subcaption}
\usepackage{booktabs}
\usepackage{mathrsfs}
 \usepackage{hyperref}
\def\BibTeX{{\rm B\kern-.05em{\sc i\kern-.025em b}\kern-.08em
    T\kern-.1667em\lower.7ex\hbox{E}\kern-.125emX}}
    \newcolumntype{Y}{>{\centering\arraybackslash}X}
\begin{document}

\title{One Shot is Enough for Sequential Infrared Small Target Segmentation
}

\author{
\IEEEauthorblockN{1\textsuperscript{st} Bingbing Dan}
\IEEEauthorblockA{\textit{Institute of Optics and Electronics} \\
\textit{Chinese Academy of Sciences}\\
Chengdu, Sichuan \\
danbingbing20@mails.ucas.ac.cn}
\and
\IEEEauthorblockN{2\textsuperscript{nd} Meihui Li, 3\textsuperscript{rd} Tao Tang}
\IEEEauthorblockA{\textit{Institute of Optics and Electronics} \\
\textit{Chinese Academy of Sciences}\\
Chengdu, Sichuan \\
limeihui@ioe.ac.cn,taotang@ioe.ac.cn}
\and
\IEEEauthorblockN{4\textsuperscript{th} Jing Zhang}
\IEEEauthorblockA{\textit{School of Computing} \\
\textit{Austrilian National University}\\
Canberra, Australia \\
jing.zhang@anu.edu.au}
}

\maketitle

\begin{abstract}
Infrared small target sequences exhibit strong similarities between frames and contain rich contextual information, which motivates us to achieve sequential infrared small target segmentation (IRSTS) with minimal data. Inspired by the success of Segment Anything Model (SAM) across various downstream tasks, we propose a one-shot and training-free method that perfectly adapts SAM's zero-shot generalization capability to sequential IRSTS. Specifically, we first obtain a confidence map through local feature matching (LFM). The highest point in the confidence map is used as the prompt to replace the manual prompt. Then, to address the over-segmentation issue caused by the domain gap, we design the point prompt-centric focusing (PPCF) module. Subsequently, to prevent miss and false detections, we introduce the triple-level ensemble (TLE) module to produce the final mask. Experiments demonstrate that our method requires only one shot to achieve comparable performance to state-of-the-art IRSTS methods and significantly outperforms other one-shot segmentation methods. Moreover, ablation studies confirm the robustness of our method in the type of annotations and the selection of reference images. Our codes are available at \href{https://github.com/D-IceIce/one-shot-IRSTS}{https://github.com/D-IceIce/one-shot-IRSTS}.

\end{abstract}


\section{Introduction}
\label{Introduction}
\IEEEPARstart{I}{nfrared} small target segmentation (IRSTS), also known as infrared small target detection (IRSTD), is an important technology in the security field with many applications~\cite{survey1,survey2}. In recent years, research on IRSTS has mainly focused on single-frame images~\cite{ACM,DNA,UIU,MSH,MCLC,LESPS}, but multi-frame sequences should not be ignored. In many practical applications~\cite{RDIAN,sequence}, detectors are often stationary or move slowly, resulting in consecutive frames that are relatively static or change slowly~\cite{slow}. Under these conditions, sequential segmentation becomes more appropriate. Therefore, considering the wide application value of sequential scenes, this paper explores sequential IRSTS.

\begin{figure}[t]
    \centering
    \includegraphics[width=0.6\linewidth]{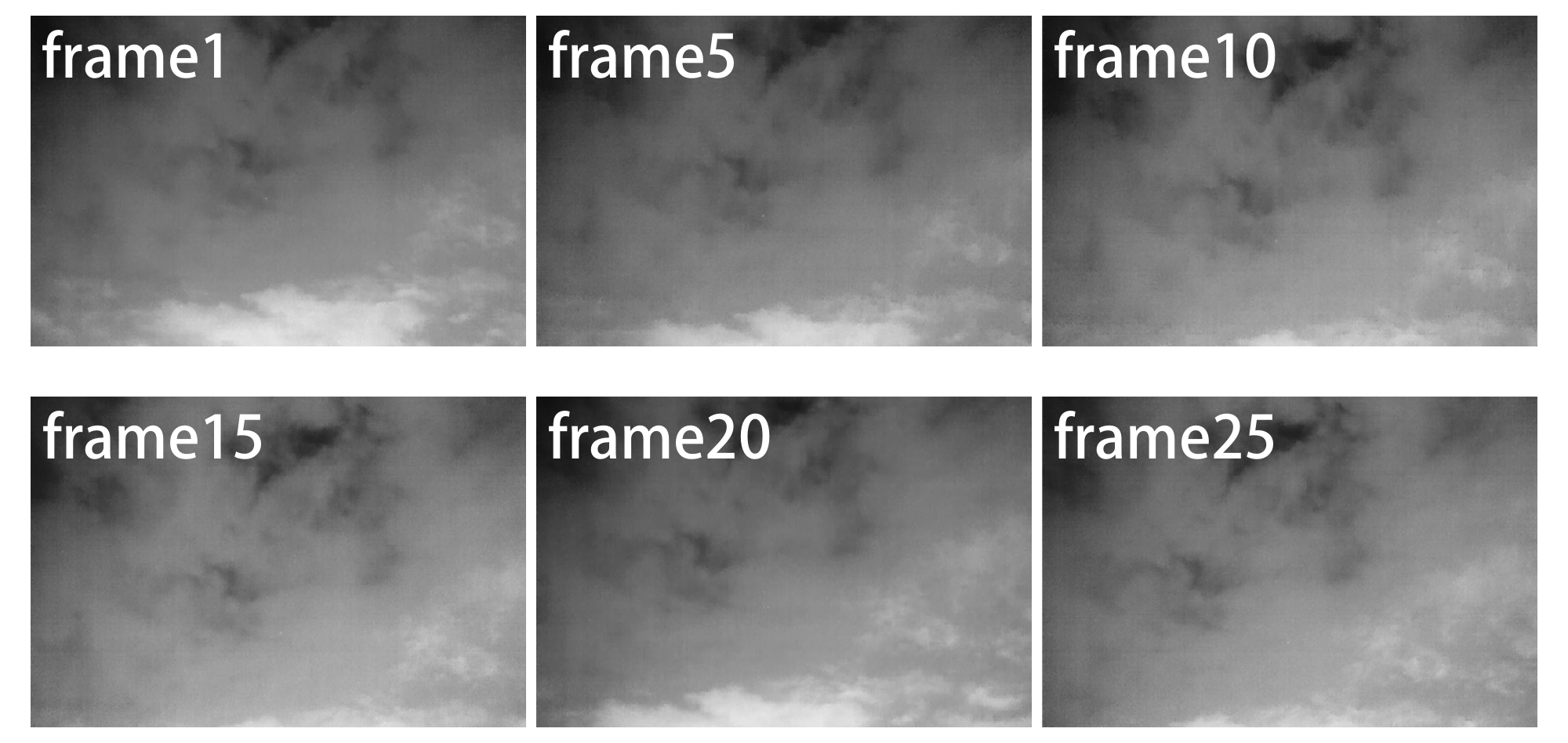}
    \includegraphics[height=0.3\linewidth]{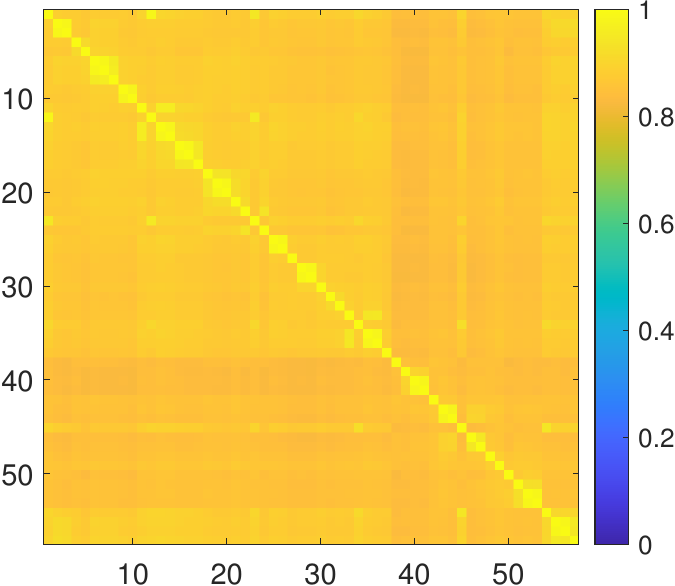}
    \caption{Characteristics of sequential small target images. Left: Images at intervals within a sequence. Right: SSIM values between each frame in the sequence.}
    \label{SeqAnalysis}
\end{figure}

To begin, we provide an analysis of the properties inherent to sequential infrared small target images, as shown in Fig. \ref{SeqAnalysis}. Due to long-distance imaging, the background changes very slowly, and there is a significant similarity between the backgrounds of different frames, as indicated by the structural similarity index measure (SSIM) between frames nearly up to 1. Concerning the targets, they consistently maintain the characteristic of being ``small" almost without change. Based on these characteristics of sequential images, we believe that the contextual information from the previous frame can be of great value for target segmentation in subsequent frames. For traditional methods requiring large amounts of training data, the learned features are often redundant. Therefore, we question whether extensive data training can be avoided and whether IRSTS in sequential scenes can be achieved with a small amount of data or even one-shot learning.

In general segmentation, some zero-shot and few-shot methods based on large pre-trained models have gained popularity, e.g. segment anything model (SAM) \cite{SAM}. These models, leveraging the rich knowledge learned from extensive data during the pre-training phase, can effectively address a wide range of perception tasks without additional training. Inspired by the successful application of large models in various downstream tasks, we believe that large models are very suitable for solving our proposed problem: achieving IRSTS in sequential scenes with a small amount of data. However, applying SAM to IRSTS presents two challenges: 1) Frame-by-frame manual prompts. For the segmentation of specific objects, SAM needs a manual prompt for each image, which is impossible for practical applications involving numerous IRSTS sequences. 2) Domain gap. During the pre-training phase, SAM focuses on RGB generic objects, resulting in insufficient accuracy when SAM directly segments tiny infrared targets.

\begin{figure*}[!ht]
\centering
\includegraphics[width=\linewidth]{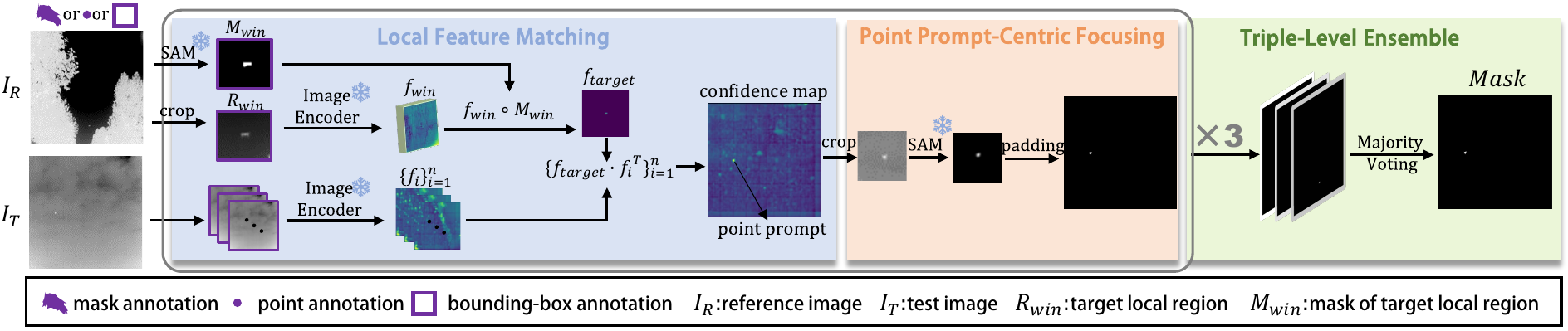}
\caption{Overview of the proposed method. It consists of three parts: local feature matching, prompt-centric focusing, and triple-level ensemble.}
\label{pipeline}
\end{figure*}

To address these challenges and achieve sequential IRSTS with a small amount of data, we propose an innovative training-free and one-shot sequential IRSTS method, as shown in Fig. \ref{pipeline}. For any given sequence, our method only requires one shot, i.e., a single frame from the sequence with annotation such as a mask, point, or bounding box, to segment small targets in other frames of the sequence. Specifically, our method consists of three steps: the first step involves local feature matching (LFM) to generate a confidence map, where we compute the similarity between given small target features in the reference image and the test image. The highest value in the confidence map is used as the point prompt to replace the manual prompt. The second step is point prompt-centric focusing (PPCF), which enables the SAM to adapt to the segmentation of tiny objects through a focus from global to local, thereby reducing the domain gap. Finally, the triple-level ensemble (TLE) module is used to merge the masks obtained from the first two steps across three different window sizes, enhancing accuracy through complementarity.

Our main contributions are summarized as follows:
\begin{itemize}
\item This is the first one-shot IRSTS method that requires no training. Only one annotated frame is provided as a reference, enabling accurate segmentation of the other frames in the sequence.
\item We fully leverage the strong generalization of SAM and introduce the LFM, PPCF, and TLE modules to address the challenges of frame-by-frame manual prompts and domain gap in the application of SAM to IRSTS.
\item Using only one shot, our method achieves comparable performance to state-of-the-art IRSTS models trained on large-scale data and significantly outperforms other one-shot segmentation methods.
\end{itemize}

\section{Proposed Method}

\subsection{Local Feature Matching}
Feature matching involves calculating the similarity between the annotated mask area in the reference image $I_R$ and the test image $I_T$, identifying small target regions in the test image $I_T$. 
If the annotation of the reference image is a point or a bounding box, we can first obtain the mask of the reference image using SAM. 

Due to the small size of the target, directly extracting global features can result in the compression or disappearance of target features. Therefore, we narrow down the area for local feature extraction. For the reference image $I_R$, a $L\times L$ window is cropped from both the reference image and mask based on the annotation center, producing $R_{\text{win}}\in \mathbb{R}^{L \times L}$ and $M_{\text{win}}\in \mathbb{R}^{L \times L}$. 

Then, the feature extraction is performed as:
\begin{equation}
\begin{aligned}
f_{\text{win}} = {\rm ImageEncoder}(R_{\text{win}})
\end{aligned}
\end{equation}
where SAM's pre-trained image encoder is utilized as a feature extractor. Next, $M_{\text{win}}$ is resized to match the size of the $f_{\text{win}}$, and the features of the target pixels $f_{\text{target}}$ are isolated by masking $f_{\text{win}}$ with $M_{\text{win}}$, as follows:
\begin{equation}
\begin{aligned}
f_{\text{target}} = f_{\text{win}}\circ M_{\text{win}}
\end{aligned}
\end{equation}
where $\circ$ denotes element-wise multiplication. Meanwhile, for each frame $I_T$ in the sequence, we crop it into patches $\{W_i\}_{i=1}^N$ to extract features, as follows:
\begin{equation}
\begin{aligned}
\{f_i\}_{i=1}^N = {\rm ImageEncoder}(\{W_i\}_{i=1}^N)
\end{aligned}
\end{equation}

After that, we calculate sub-confidence maps by the cosine similarity between $f_{\text{target}}$ and test image feature $\{f_i\}_{i=1}^N$, and reconstruct the sub-confidence maps into a complete confidence map as follows:
\begin{equation}
\begin{aligned}
\{c_i\}_{i=1}^N = \{f_{\text{target}} \cdot f_i^T\}_{i=1}^N\\
C = \text{Reconstruct}(\{c_i\}_{i=1}^N)
\end{aligned}
\end{equation}

The confidence map $C$ indicates the degree of similarity between the features of the target and those of the corresponding regions in the test image. Higher values in $C$ correspond to regions that are more likely to match the annotated target area in the reference image. We select the point with the highest confidence value, denoted as $P$, as the point prompt, which represents the most likely center position of the small target (The issue of the highest value point not necessarily representing the target will be addressed in section C). This automatic prompt extraction fulfills the prompt input requirement for SAM, addressing the challenge of frame-by-frame manual prompts.

\subsection{Point Prompt-Centric Focusing}
\begin{figure}[t]
\centering
\includegraphics[width=\linewidth]{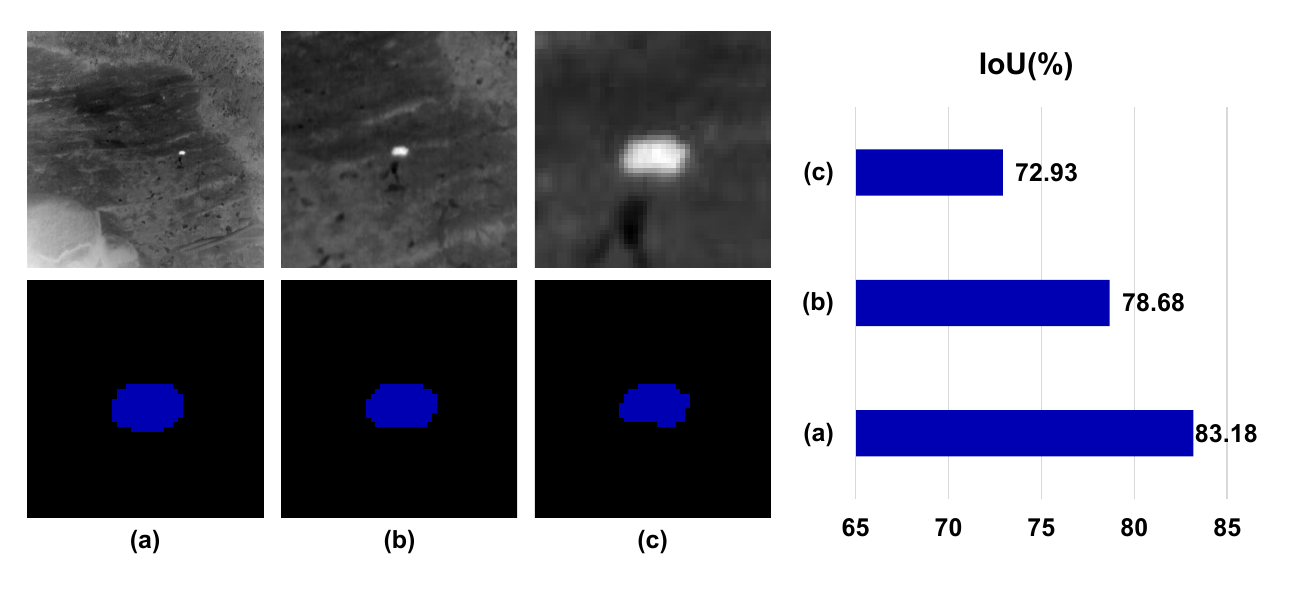}
\caption{The effect of the size of the target object relative to the image on segmentation accuracy of SAM.}
\label{differentScale}
\end{figure}

Ideally, when an image with a point prompt is fed into SAM, SAM tends to segment the contiguous region surrounding the point prompt. However, due to the domain gap, SAM may over-segment small targets with blurry boundaries, resulting in the segmentation mask including all possible surrounding areas, shown as Fig. \ref{differentScale}(a). Interestingly, we have discovered a phenomenon that helps SAM segment small targets more accurately, which is that the size of the target object relative to the image can affect the segmentation accuracy of SAM. As shown in Fig. \ref{differentScale}, we crop an infrared small target image at different sizes, each with the center point of the target area as the prompt. After segmenting with SAM, the masks obtained have varying $IoU$ values. A larger region may result in a coarser segmentation mask and lower $IoU$ due to the inclusion of more background information, which can dilute the focus on the target object. Conversely, a smaller region, tightly centered around the object, typically yields a more precise segmentation mask and higher $IoU$. This finding highlights the importance of narrow region size when using SAM to achieve precise IRSTS. 

Therefore, we propose the point prompt-centric focusing (PPCF) module to address the over-segment of small targets by focusing the image from the global view to a local region centered on the point prompt. Specifically, a $L\times L$ local window $I_l$ is cropped around the point prompt, and the coordinates of the point prompt are adjusted accordingly $P \to P_l$. Then, both $I_l$ and $P'$ are fed into SAM to obtain a local mask $M_l$. The local mask $M_l$ is then integrated back into the corresponding location of the original reference image to produce the segmentation mask $M$ for the entire image. The entire process is as follows:
\begin{equation}
\begin{aligned}
& I_l = I_R[P_x-\frac{L}{2}:P_x+\frac{L}{2}, P_y-\frac{L}{2}:P_y+\frac{L}{2}]\\
& P_l = (\frac{L}{2}+1, \frac{L}{2}+1)\\
& M_l = \text{SAM}(I_l, P_l)\\
& M[P_x-L:P_x+L,P_y-L:P_y+L] = M_l
\end{aligned}
\end{equation}

This PPCF process ensures precise segmentation and higher quality compared to direct segmentation with SAM.

\subsection{Triple-Level Ensemble}
The above two steps can achieve preliminary segmentation of small targets, both involving the concept of ``window", where the window size significantly influences these steps. Using a larger window to calculate feature similarity can lead to false alarms, resulting in the highest value point not necessarily being the true target, but rather strong edges or corners. When the window is too large, the target features may even be submerged by background features, leading to unsuccessful target matching. Conversely, using a smaller window focuses excessively on local details, and even minor differences between the reference image and the test image can result in the target not being matched. Therefore, setting a single window size can limit the accuracy of the segmentation. 

We propose a triple-level ensemble (TLE) module, employing three different window sizes $L = [\frac{H}{2},\frac{H}{3},\frac{H}{4}]$ to perform the calculations for the aforementioned two steps simultaneously, resulting in three masks. These masks are then combined using a ``Majority Voting" method to ensemble into the final mask. For each pixel, the value (either target or background) that appears most frequently among the three masks is selected for the final mask, the process is as follows:
\begin{equation}
\begin{aligned}
M(x,y) = \text{mode}(M_1(x,y),M_2(x,y),M_3(x,y))
\end{aligned}
\end{equation}
where $\text{mode}$ is a statistical function that returns the most frequent value among its arguments.

The TLE module enhances the precision of the segmentation process by complementing LFM and PPCF across triple levels, thereby avoiding misses and false detections.

\section{Experiments}
\subsection{Experimental Settings}
We conduct experiments on the 85 real-sequence scenarios contained in the IRDST dataset \cite{RDIAN} and report the average results across all sequences. We utilize Intersection over Union ($IoU$, in units of $10^{-2}$) and the false alarm rate ($F_a$, in units of $10^{-6}$) as pixel-level evaluation metrics, along with the probability of detection ($P_d$, in units of $10^{-2}$) for assessing target-level performance.

\subsection{Comparison with Existing Methods}
\begin{table}[!ht]
\centering
\caption{Comparision with different types of existing segmentation models. The optimal are in \textbf{bold} and the second-best are \underline{underlined}.}
\renewcommand\arraystretch{1}{
\begin{tabularx}{\linewidth}{Yl ccc}
\midrule
Type & Method & $P_d(\uparrow)$ & $F_a(\downarrow)$ & $IoU(\uparrow)$\\
\midrule
\multirow{4}*{\makecell[c]{Many-Shot\\Conventional\\Supervised Learning}} & DNANet \cite{DNA} & 93.25 & 85.32 & 50.15\\
& UIUNet \cite{UIU} & 95.19 & 46.85 & \bf{76.86}\\
& RDIAN \cite{RDIAN} & \bf{96.53} & \underline{33.56} & 67.88\\
& MSHNet \cite{MSH} & 94.44 & 77.41 & \underline{70.19}\\
\midrule
\multirow{2}*{\makecell[c]{One-Shot\\In-context Learning}}& Painter \cite{Painter} & 5.162 & 17112 & 0.002\\
& SegGPT \cite{seggpt} & 31.79 & 149.7 & 1.08\\
\midrule
\multirow{3}*{\makecell[c]{One-Shot\\Training-Free}}& Matcher \cite{Matcher} & 49.34 & 9036 & 0.98\\
& PerSAM \cite{PerSAM} & 5.562 & 3942 & 4.05\\
& Ours & \underline{95.43} & \bf{31.02} & 49.95\\
\midrule
\end{tabularx}
}
\label{quantitative_comparison}
\end{table}

\begin{figure*}[!ht]
\centering
\begin{minipage}{0.09\linewidth}
    \includegraphics[width=\linewidth]{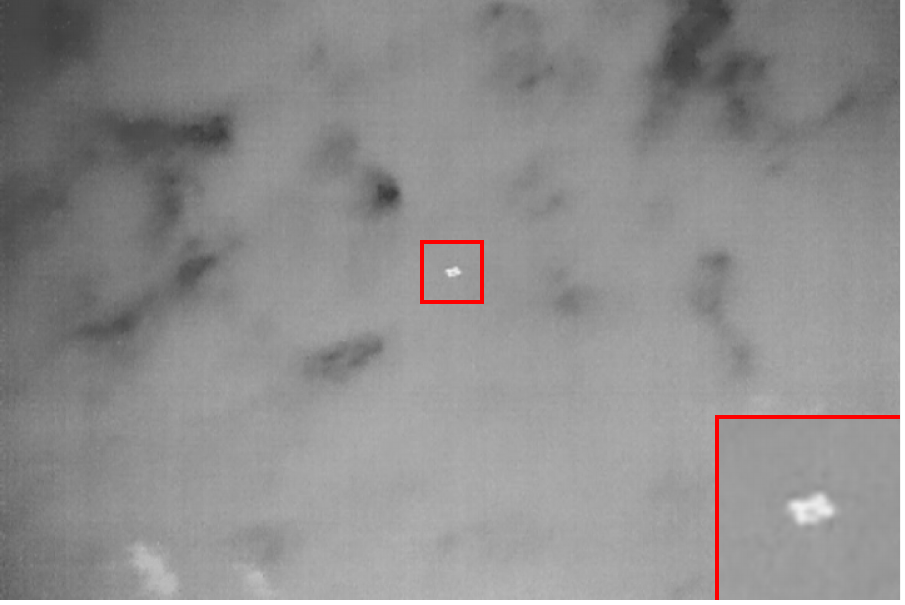}
    \includegraphics[width=\linewidth]{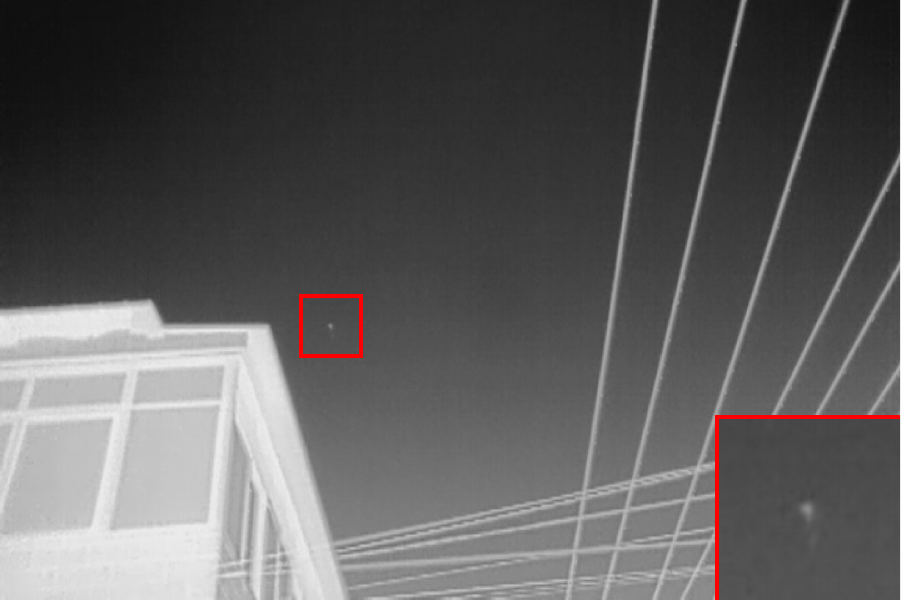}
    \includegraphics[width=\linewidth]{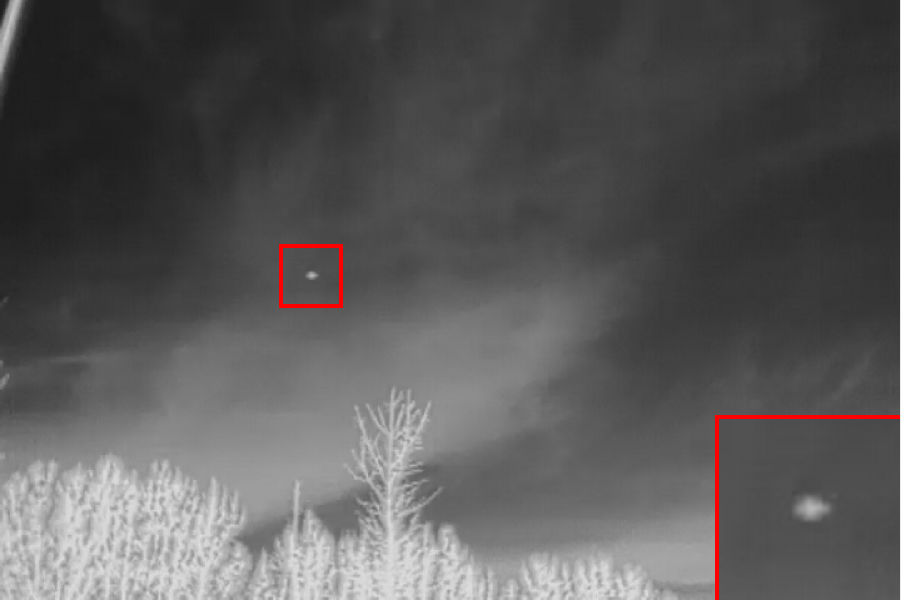}
    \includegraphics[width=\linewidth]{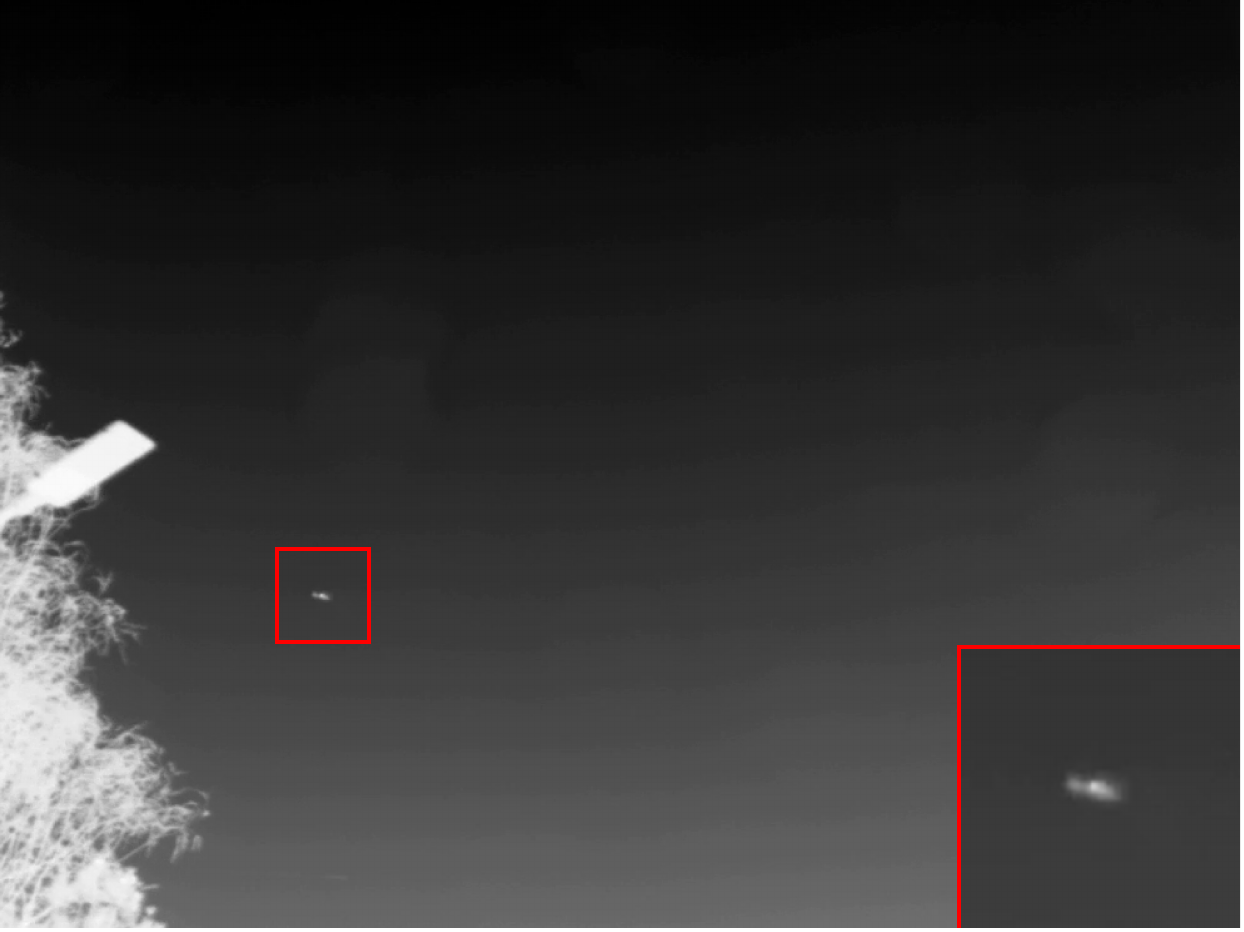}
    \centerline{Infrared Image}
\end{minipage}
\begin{minipage}{0.09\linewidth}
    \includegraphics[width=\linewidth]{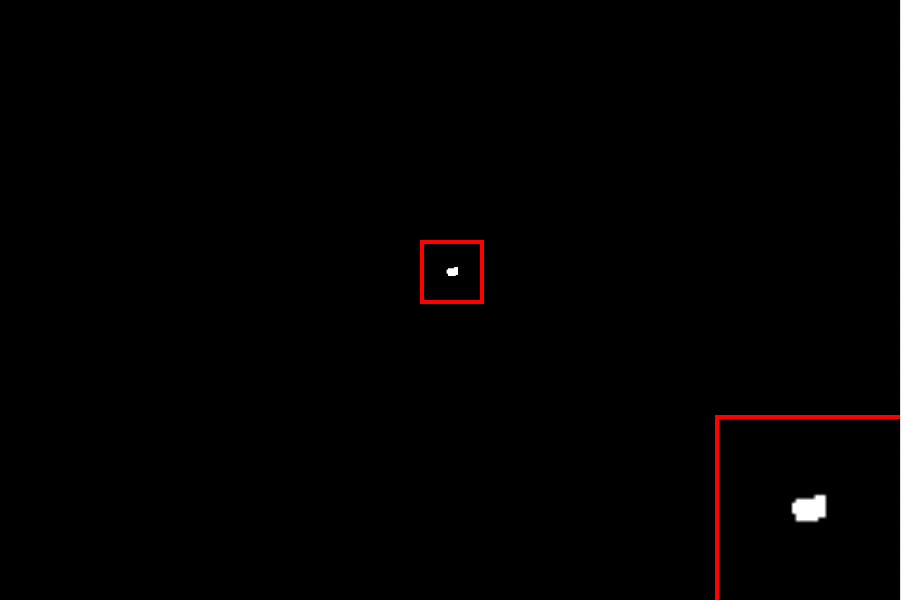}
    \includegraphics[width=\linewidth]{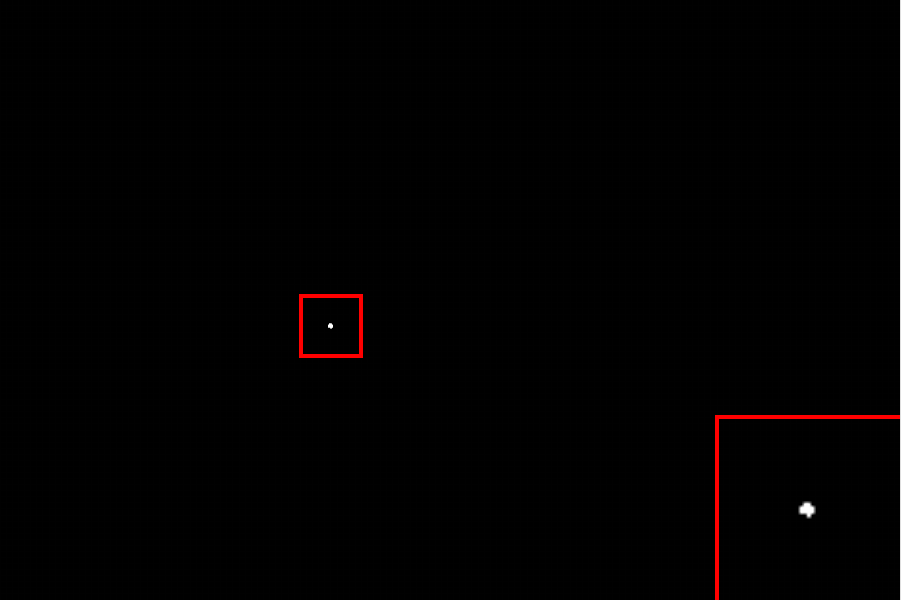}
    \includegraphics[width=\linewidth]{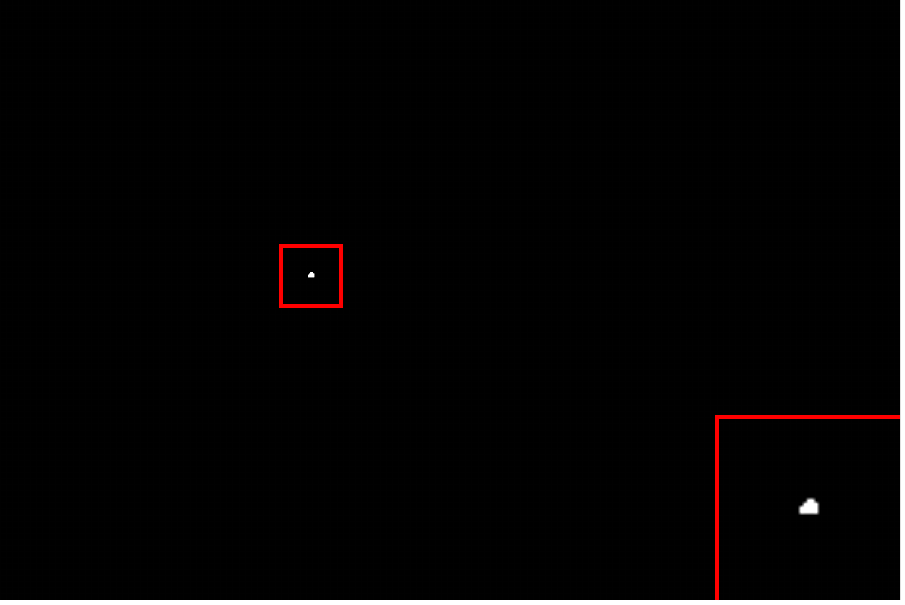}
    \includegraphics[width=\linewidth]{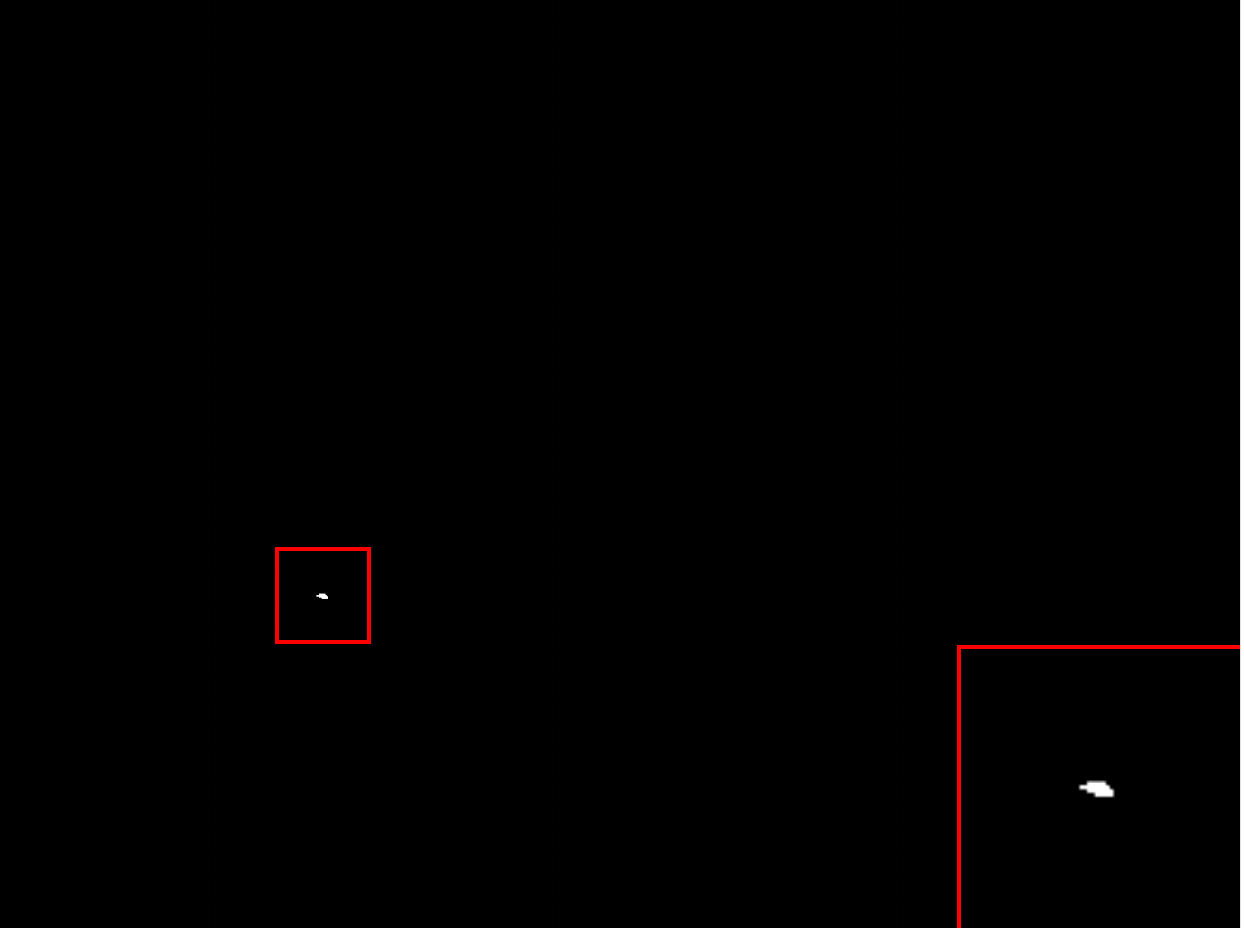}
    \centerline{GT}
\end{minipage}
\begin{minipage}{0.09\linewidth}
    \includegraphics[width=\linewidth]{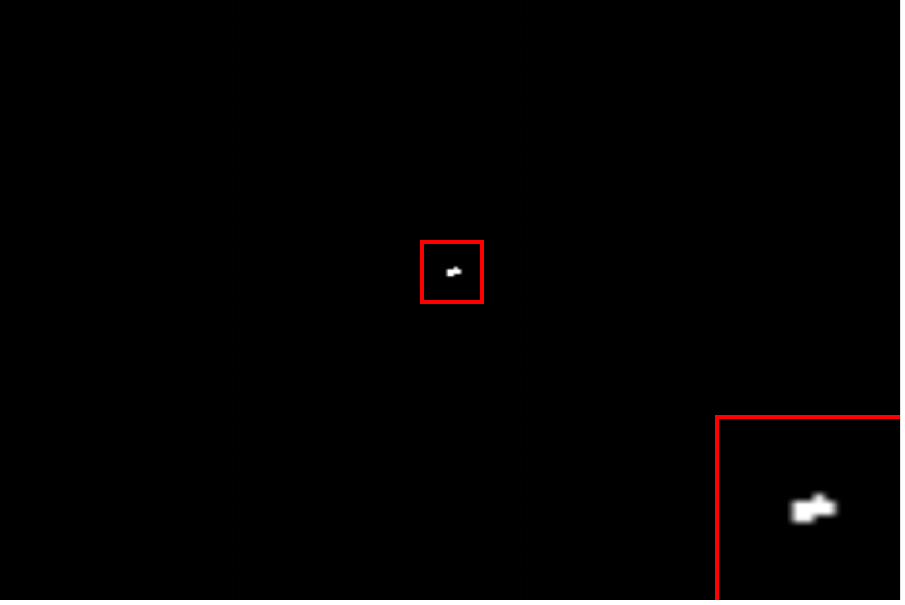}
    \includegraphics[width=\linewidth]{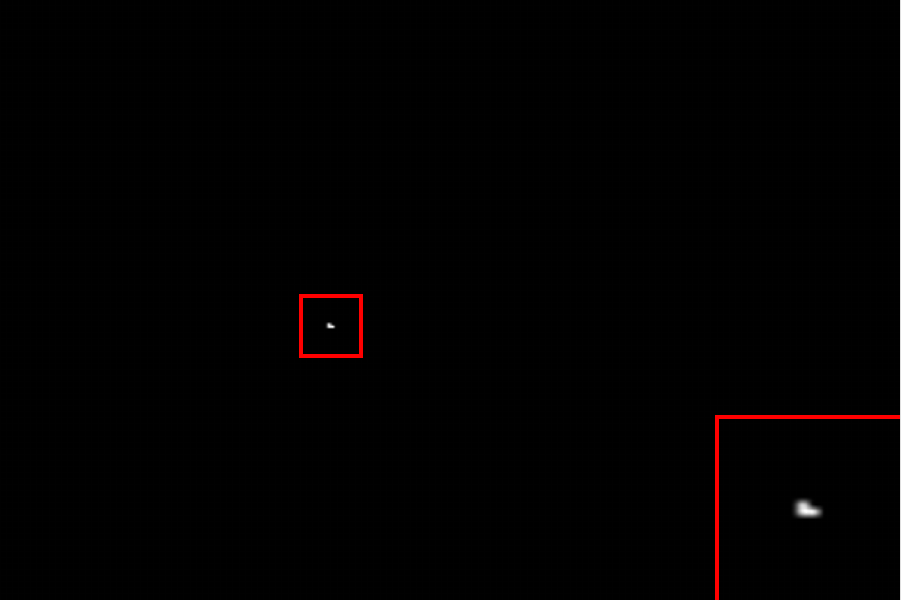}
    \includegraphics[width=\linewidth]{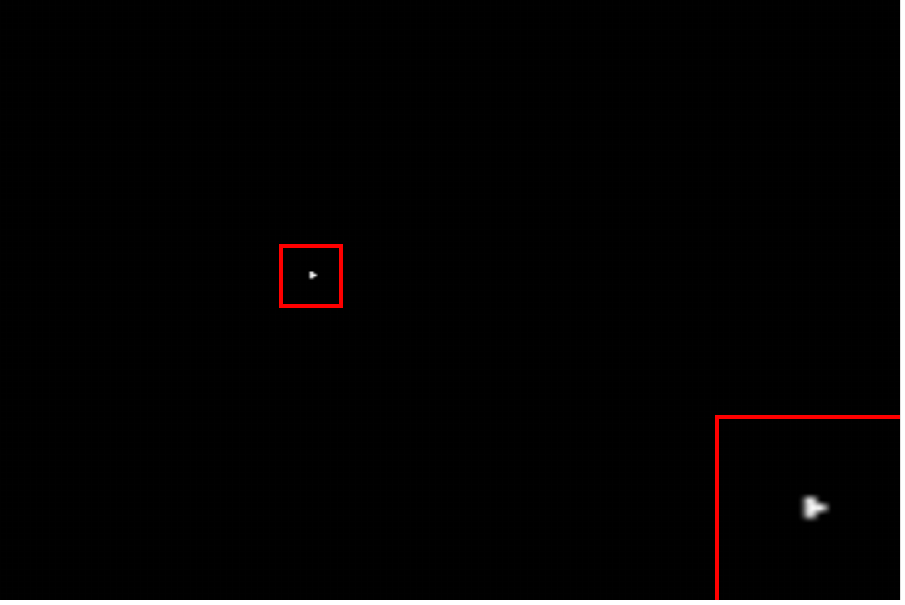}
    \includegraphics[width=\linewidth]{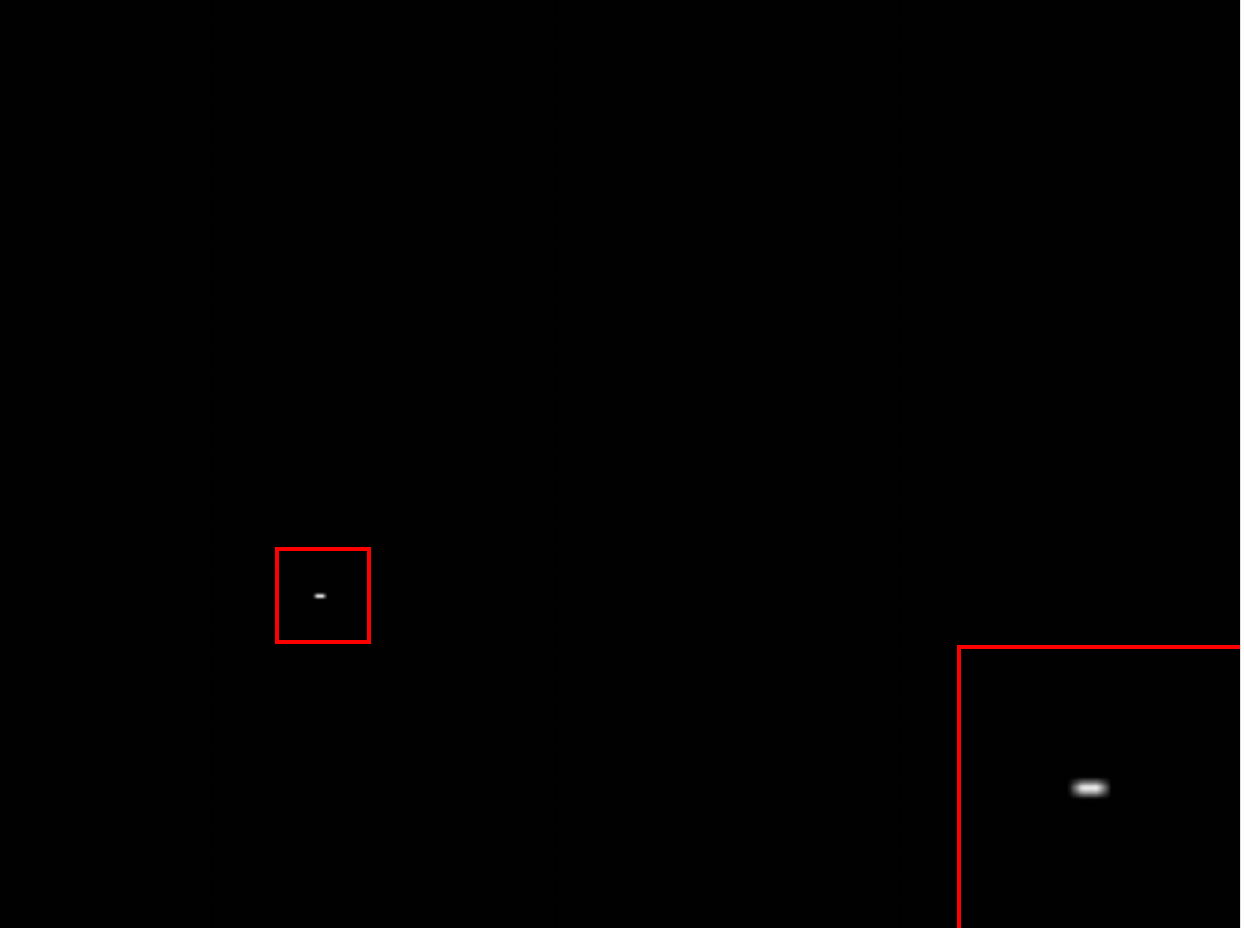}
    \centerline{DNANet}
\end{minipage}
\begin{minipage}{0.09\linewidth}
    \includegraphics[width=\linewidth]{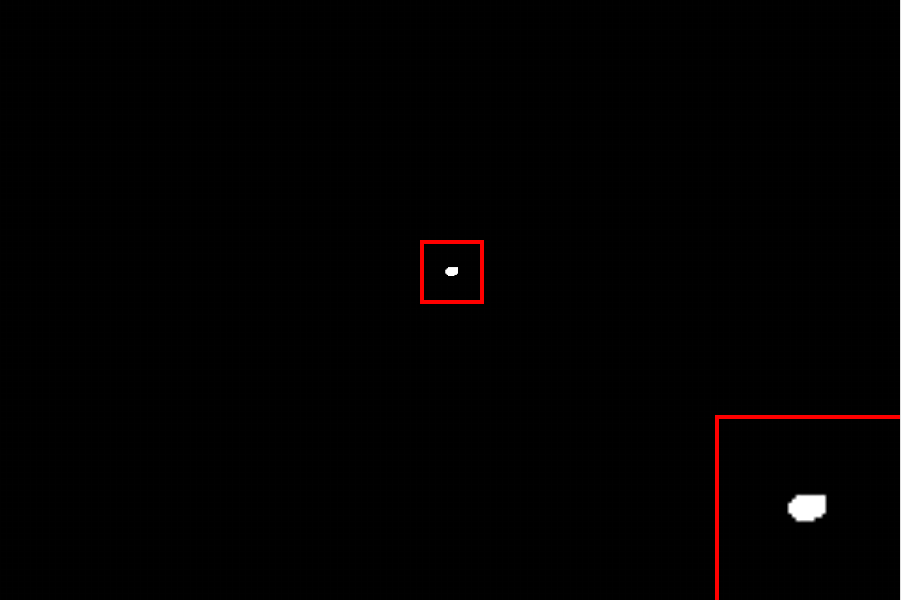}
    \includegraphics[width=\linewidth]{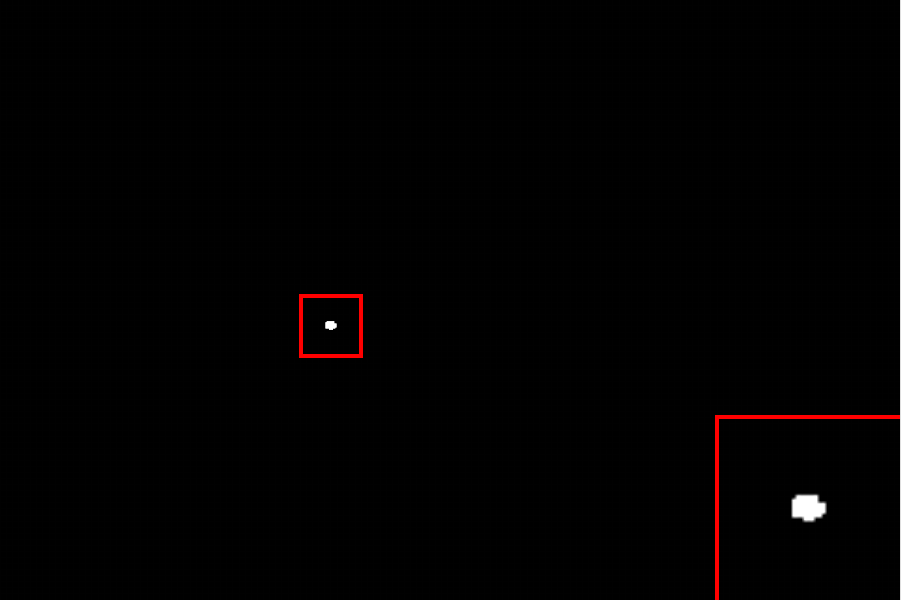}
    \includegraphics[width=\linewidth]{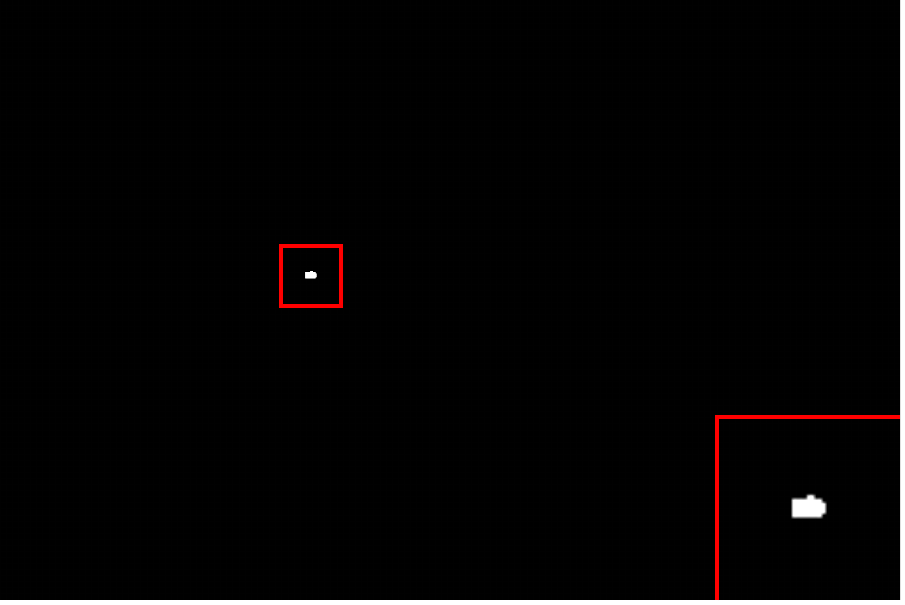}
    \includegraphics[width=\linewidth]{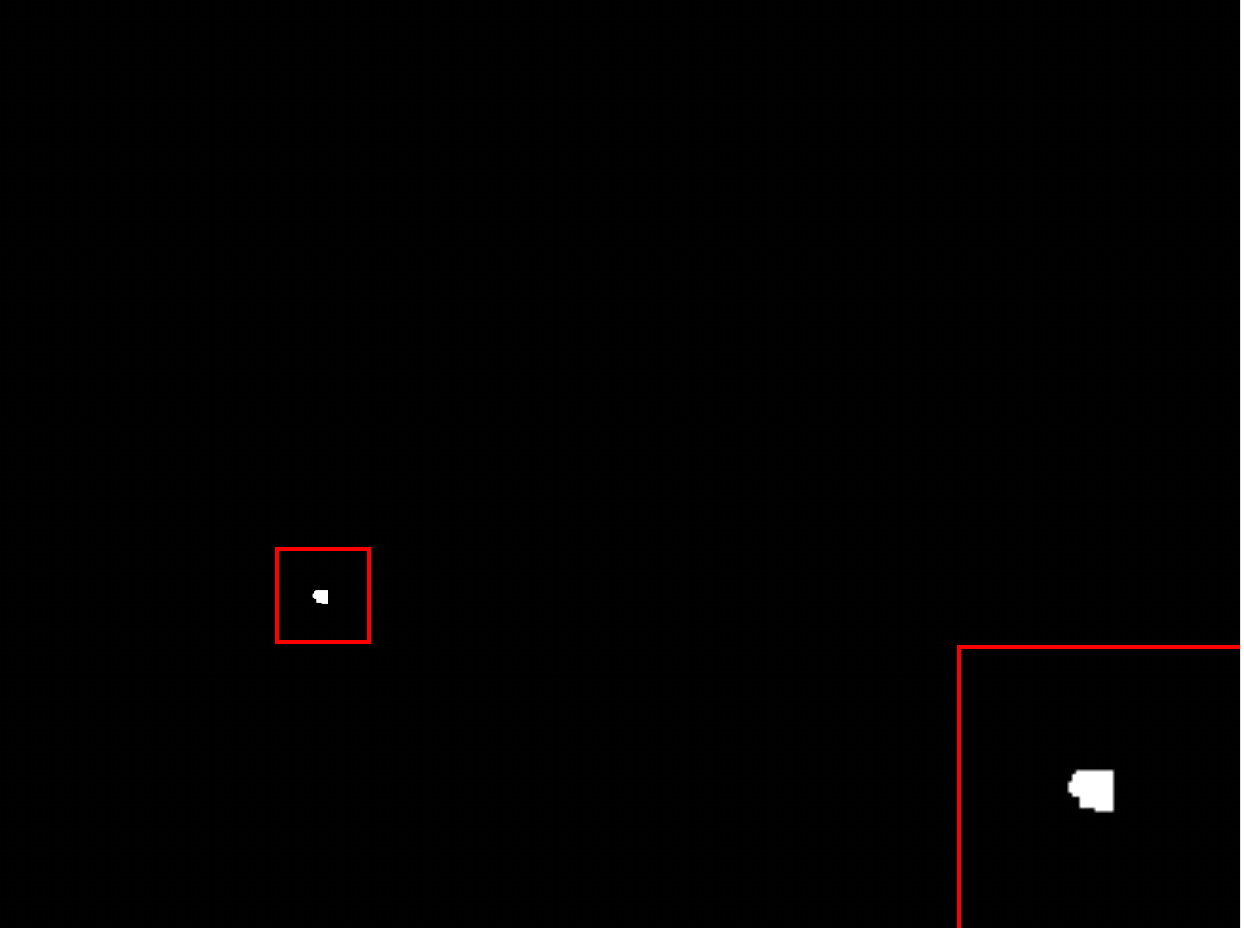}
    \centerline{UIUNet}
\end{minipage}
\begin{minipage}{0.09\linewidth}
    \includegraphics[width=\linewidth]{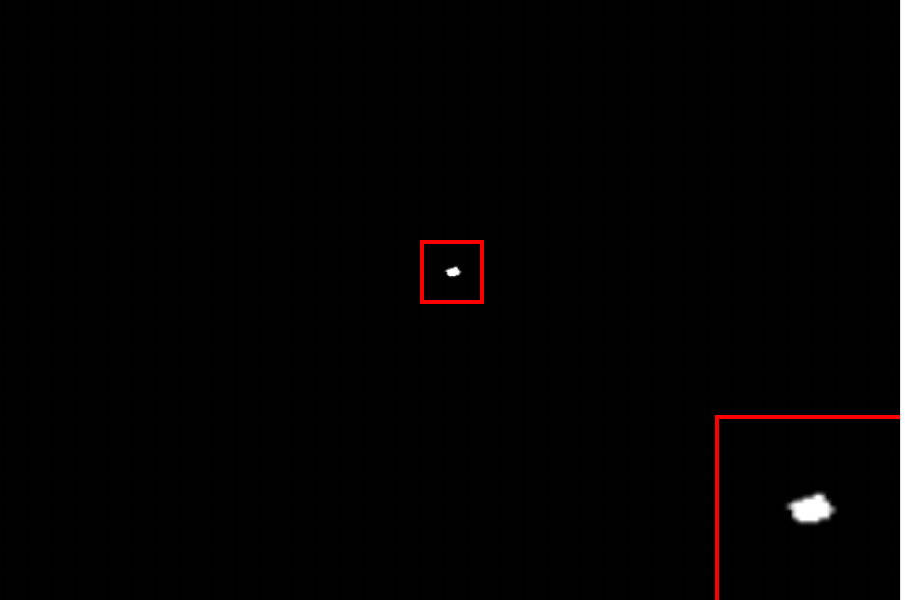}
    \includegraphics[width=\linewidth]{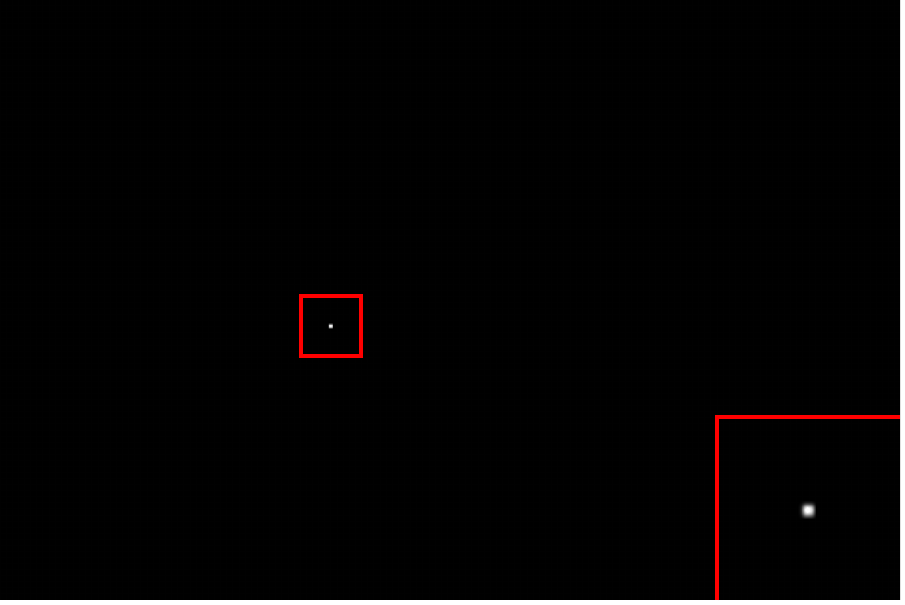}
    \includegraphics[width=\linewidth]{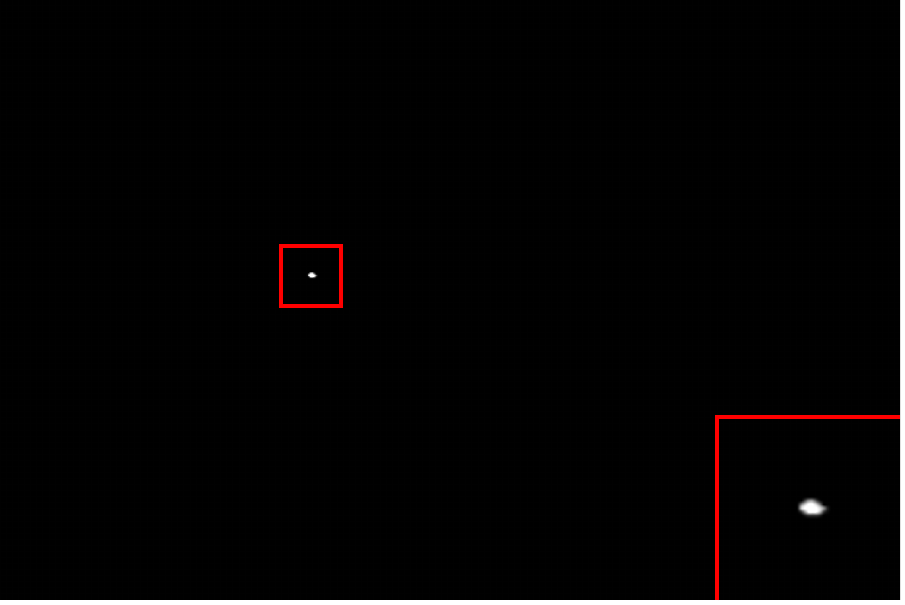}
    \includegraphics[width=\linewidth]{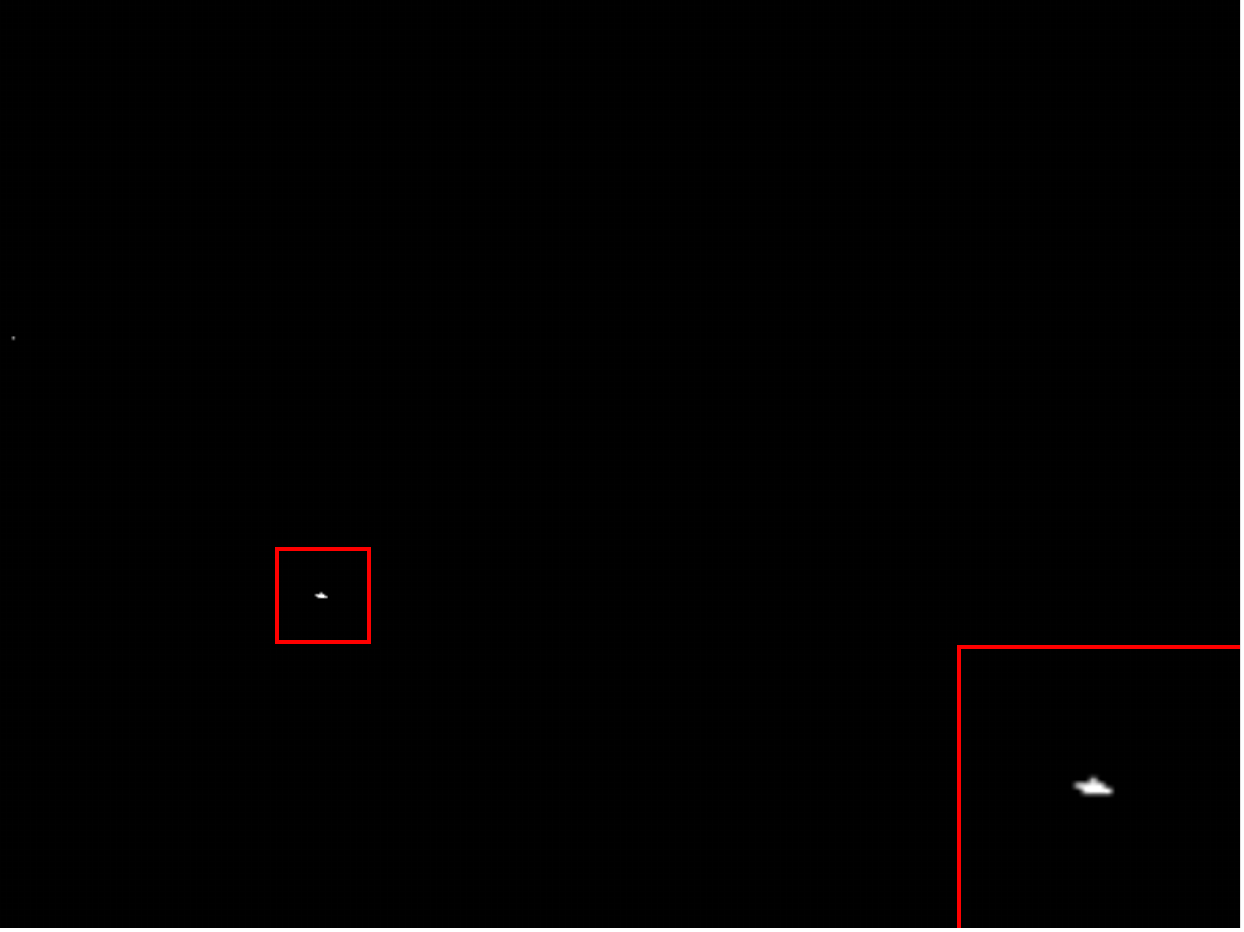}
    \centerline{RDIAN}
\end{minipage}
\begin{minipage}{0.09\linewidth}
    \includegraphics[width=\linewidth]{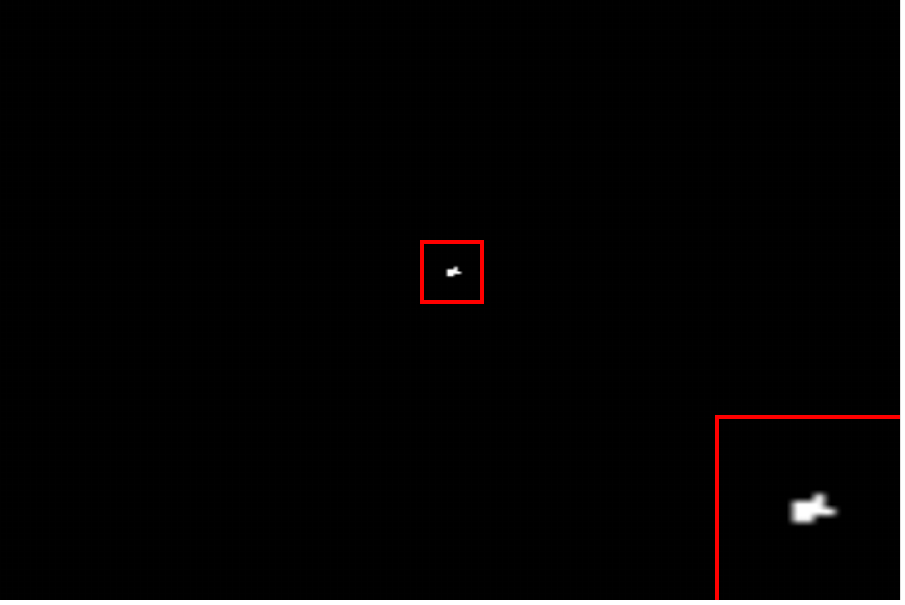}
    \includegraphics[width=\linewidth]{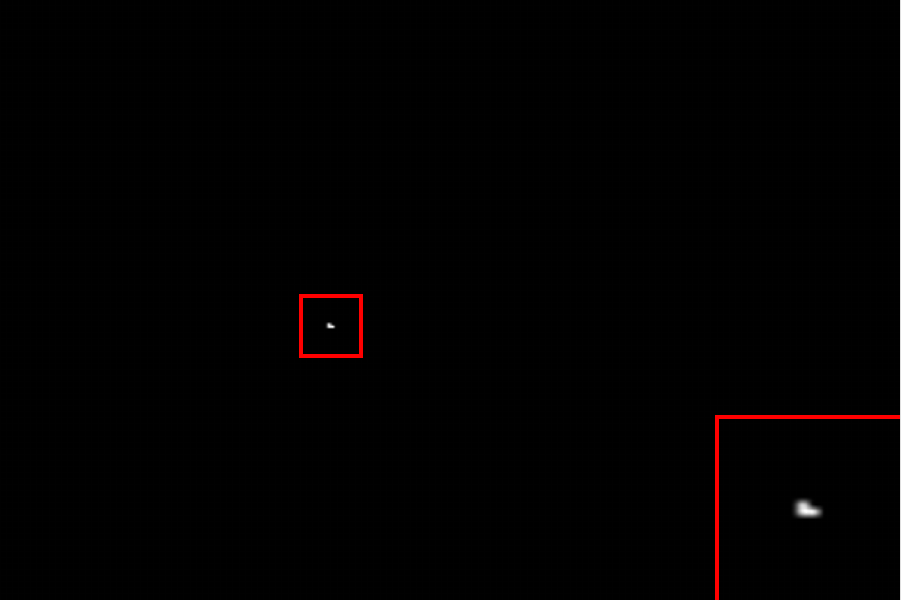}
    \includegraphics[width=\linewidth]{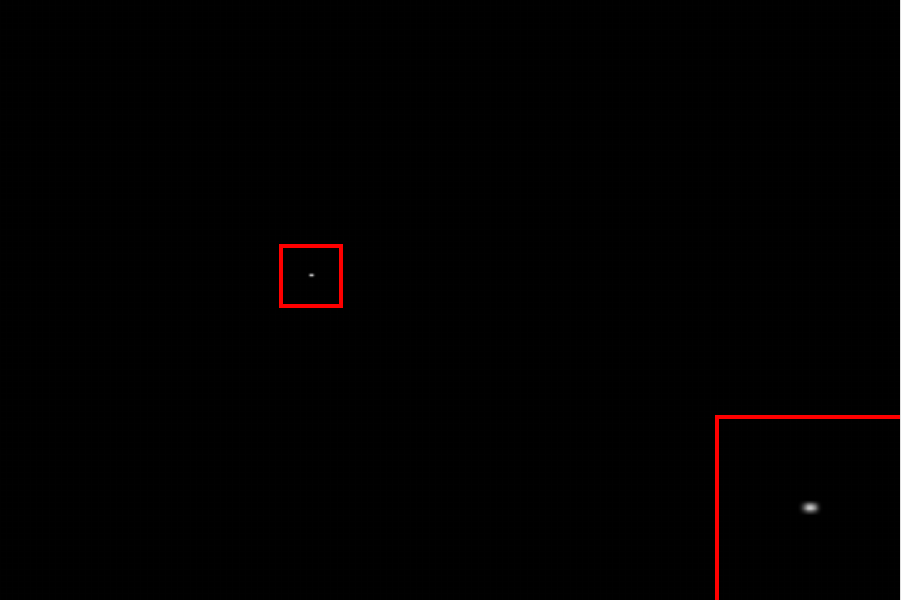}
    \includegraphics[width=\linewidth]{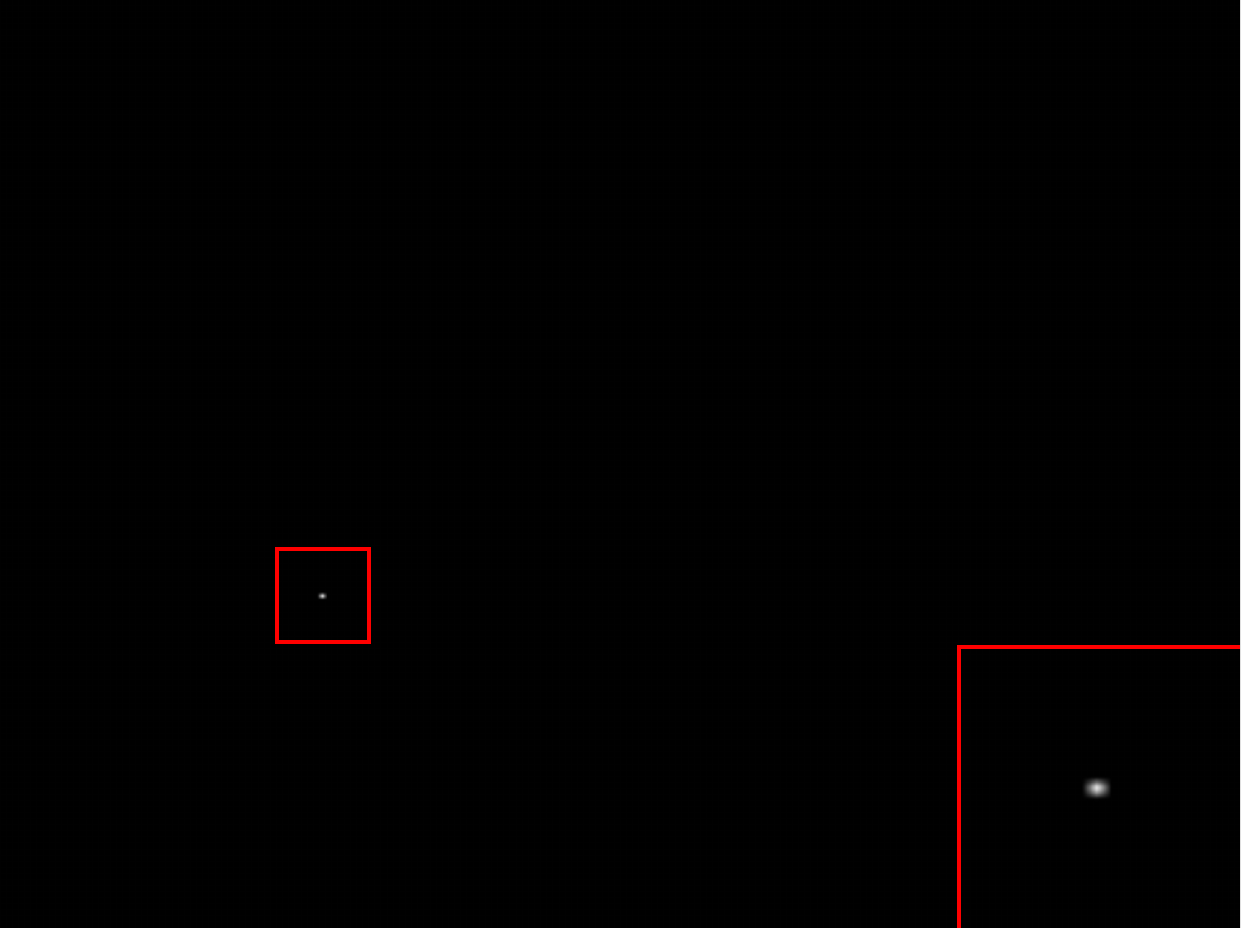}
    \centerline{MSHNet}
\end{minipage}
\begin{minipage}{0.09\linewidth}
    \includegraphics[width=\linewidth]{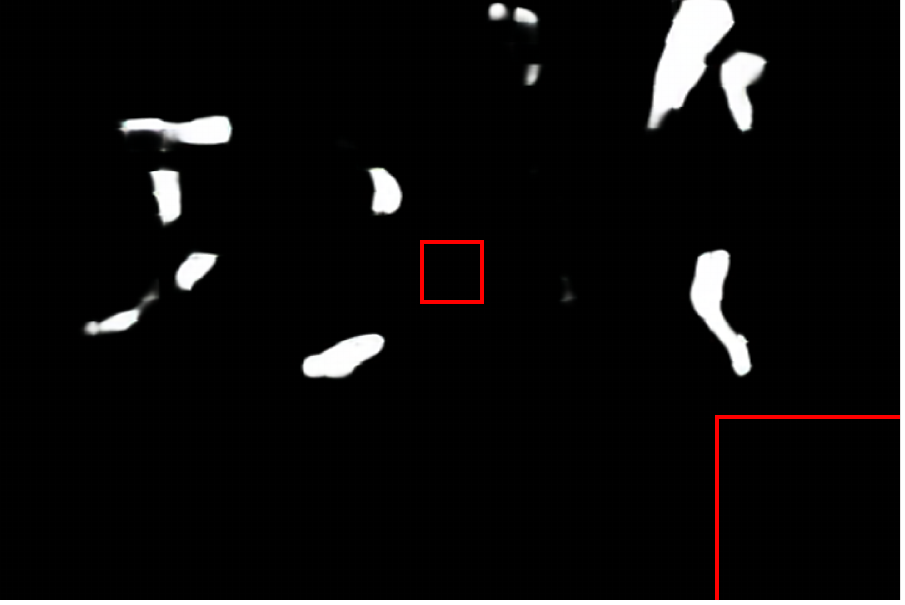}
    \includegraphics[width=\linewidth]{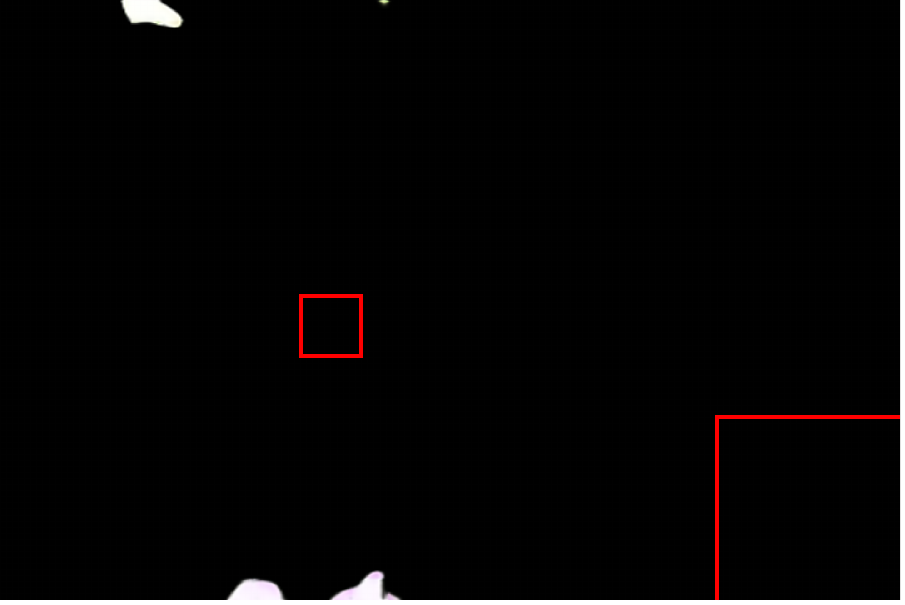}
    \includegraphics[width=\linewidth]{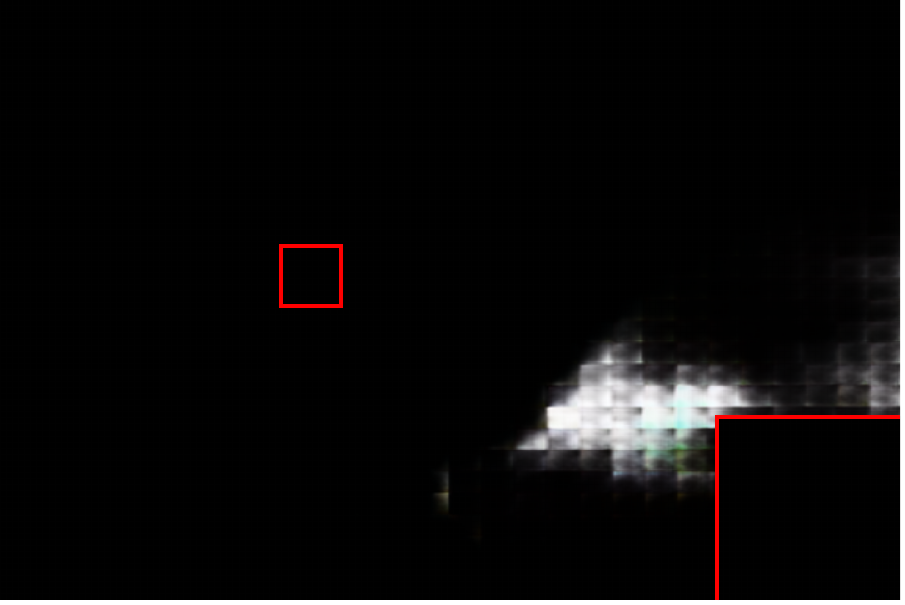}
    \includegraphics[width=\linewidth]{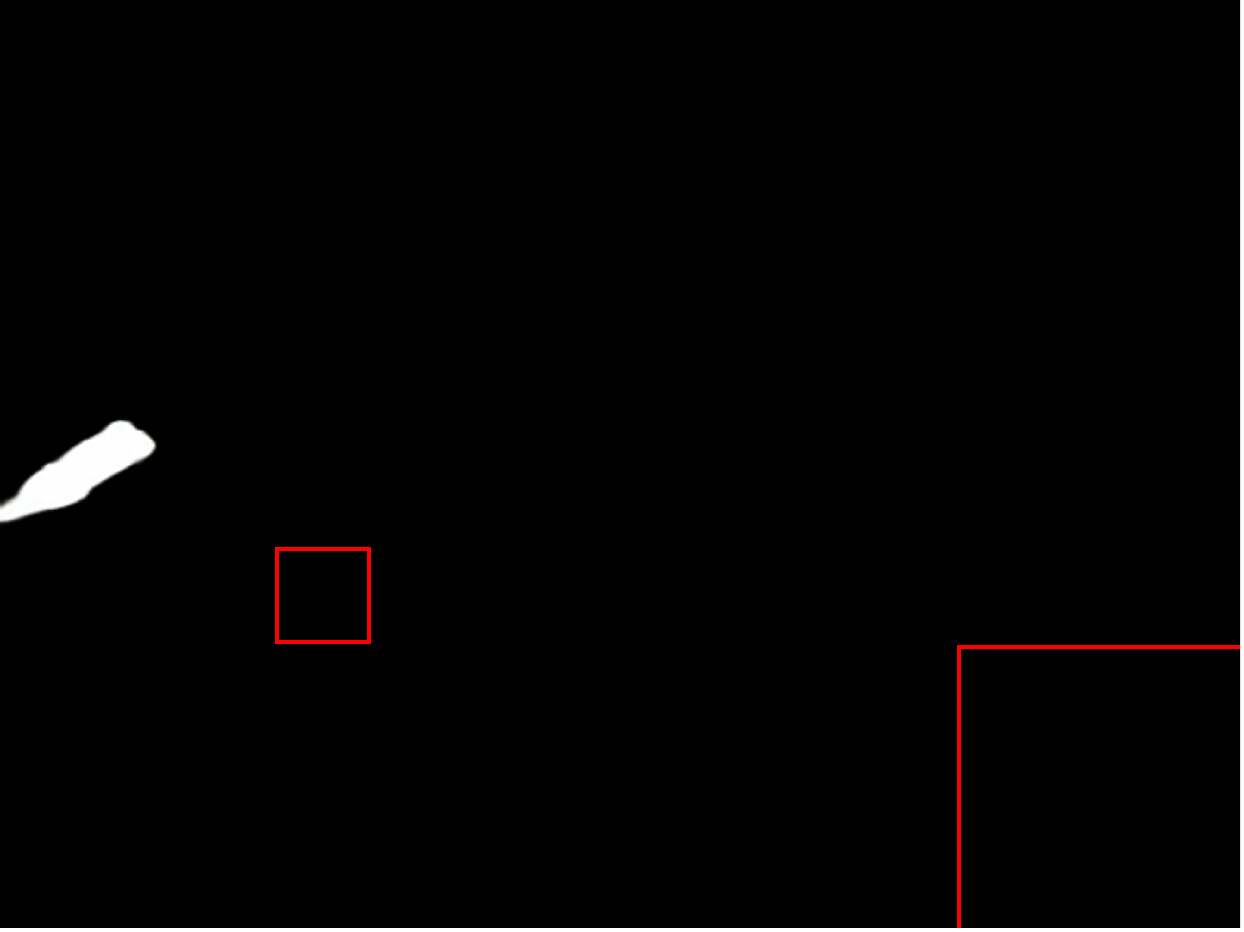}
    \centerline{Painter}
\end{minipage}
\begin{minipage}{0.09\linewidth}
    \includegraphics[width=\linewidth]{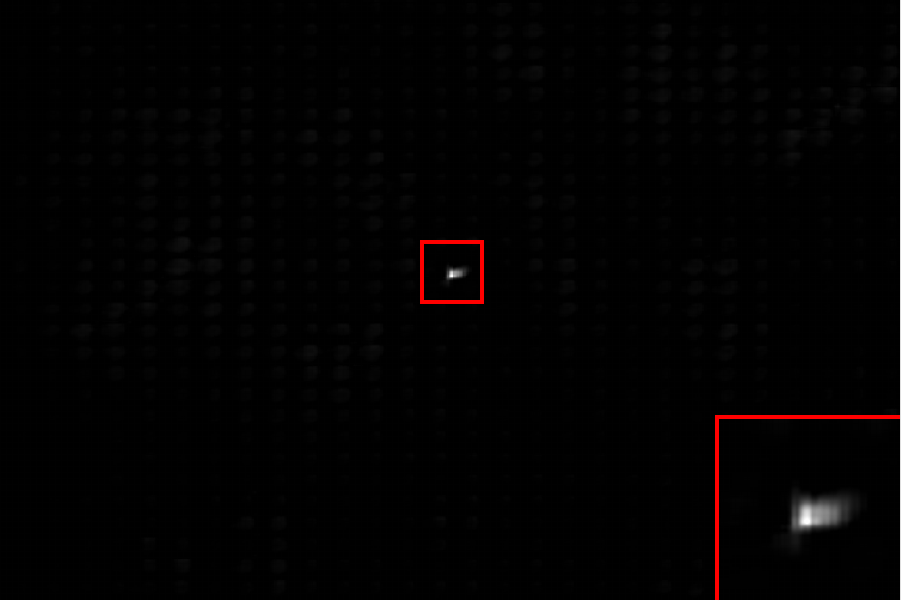}
    \includegraphics[width=\linewidth]{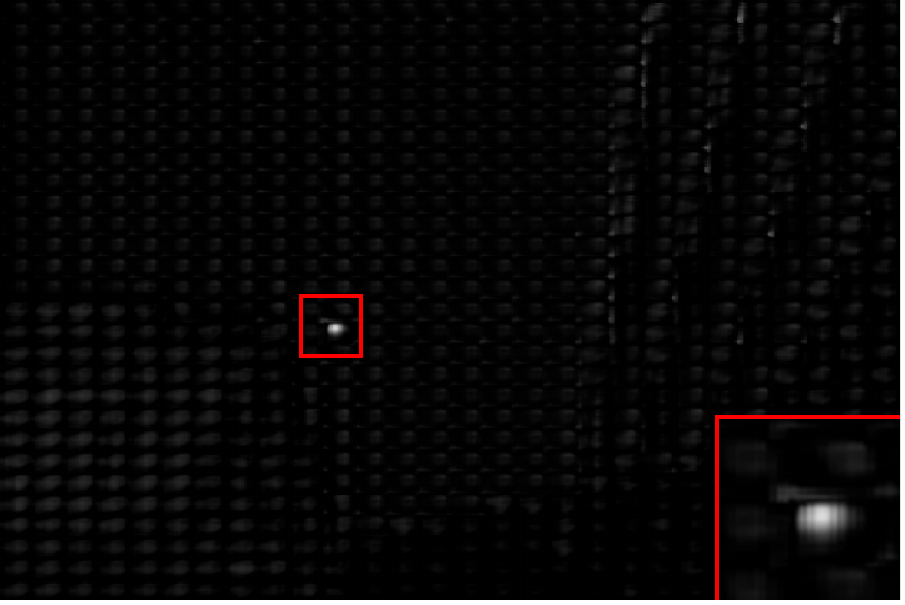}
    \includegraphics[width=\linewidth]{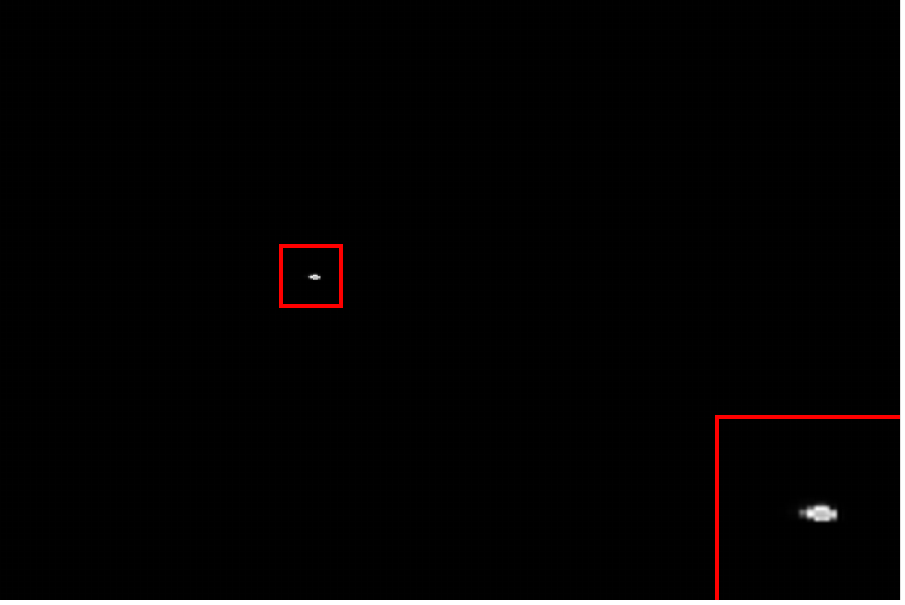}
    \includegraphics[width=\linewidth]{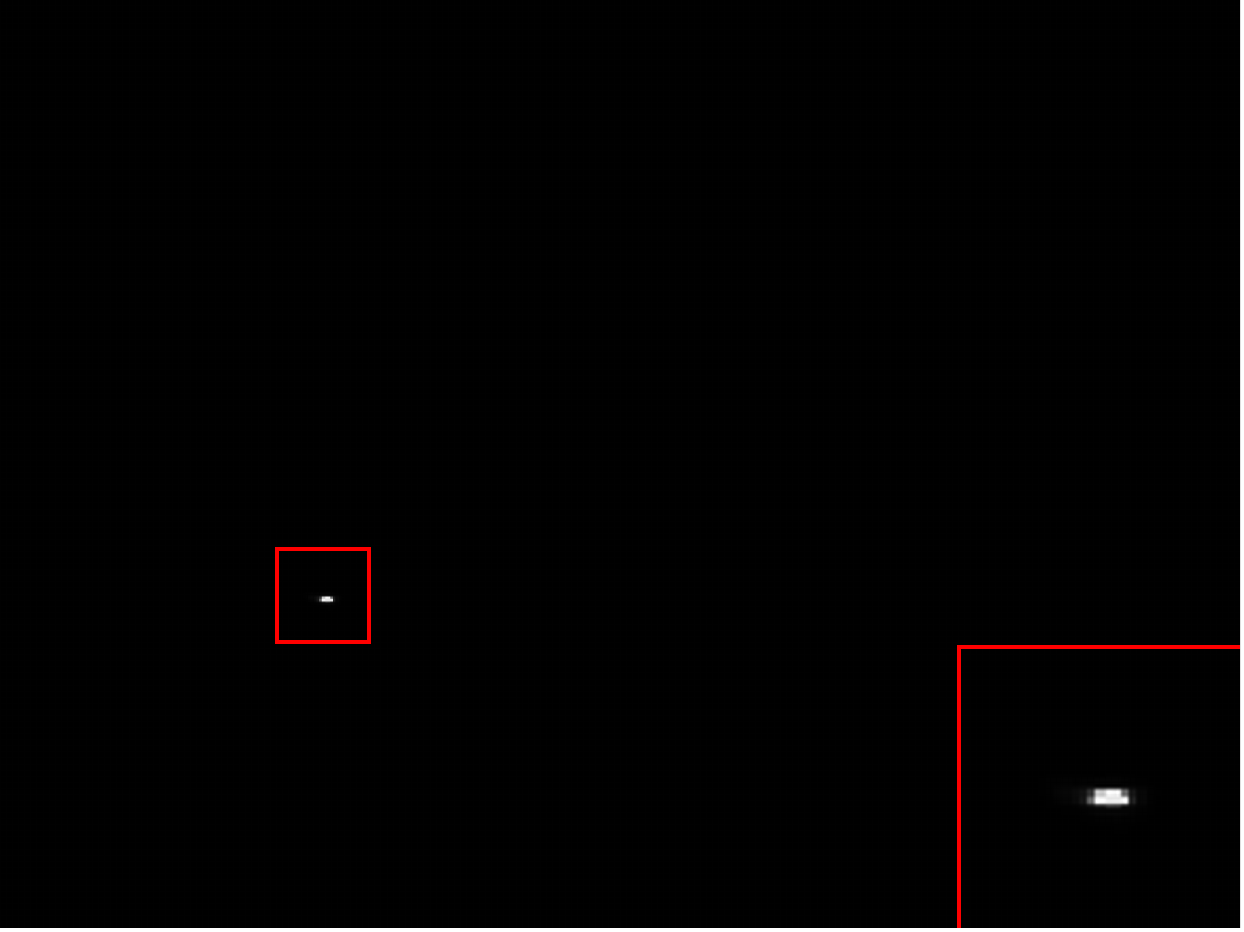}
    \centerline{SegGPT}
\end{minipage}
\begin{minipage}{0.09\linewidth}
    \includegraphics[width=\linewidth]{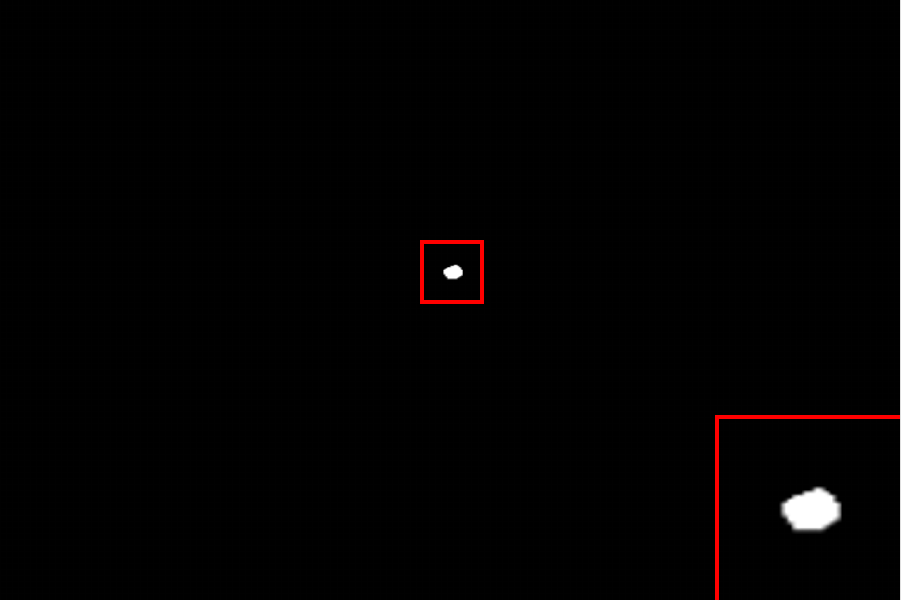}
    \includegraphics[width=\linewidth]{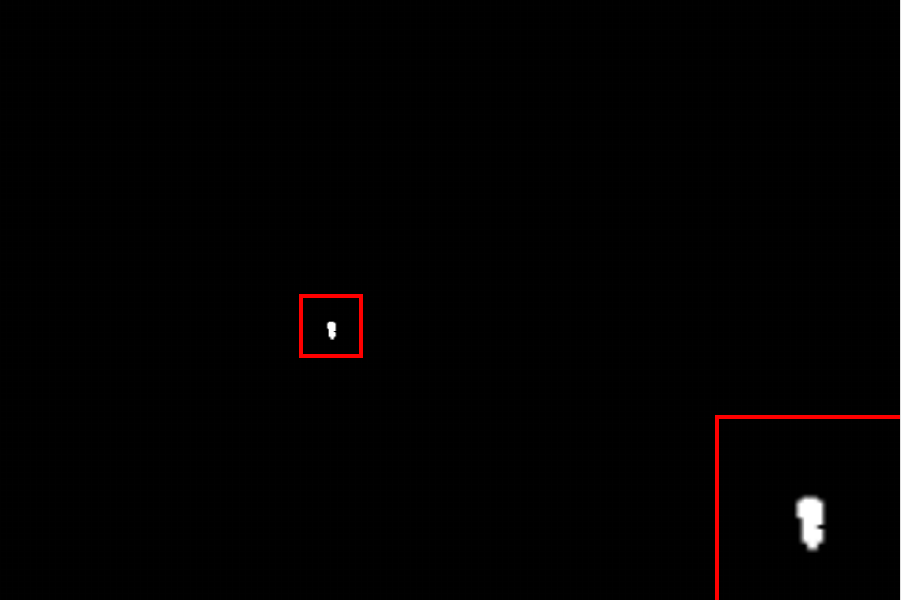}
    \includegraphics[width=\linewidth]{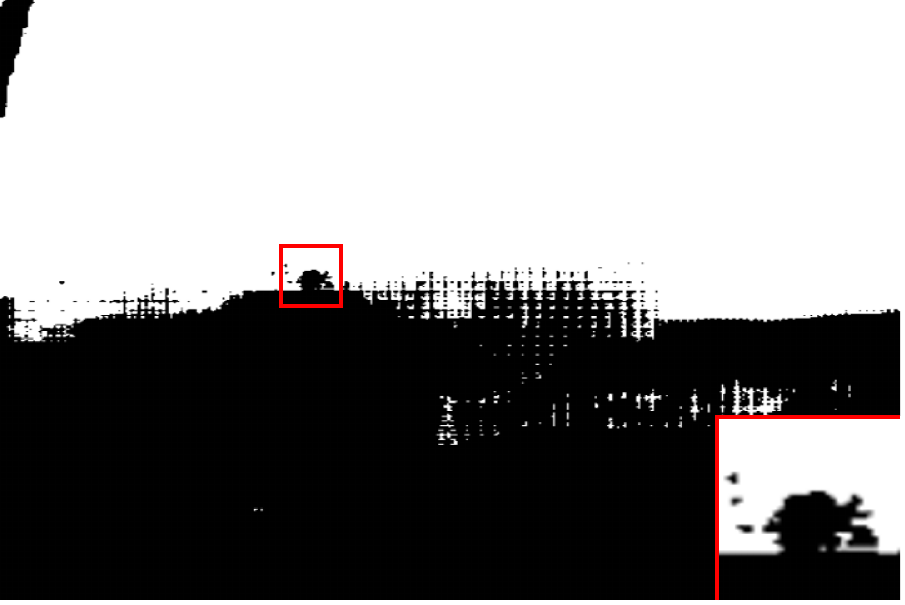}
    \includegraphics[width=\linewidth]{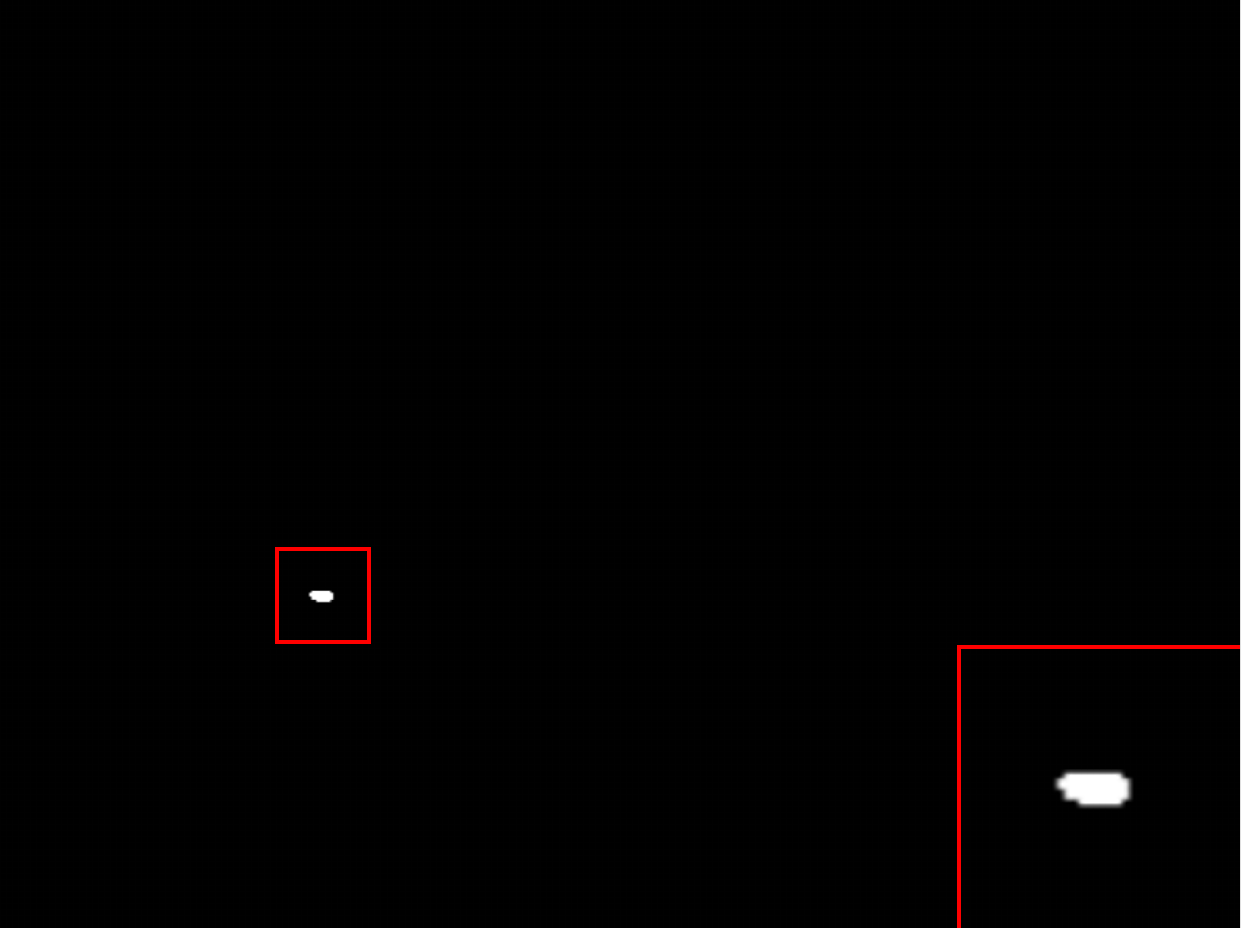}
    \centerline{Matcher}
\end{minipage}
\begin{minipage}{0.09\linewidth}
    \includegraphics[width=\linewidth]{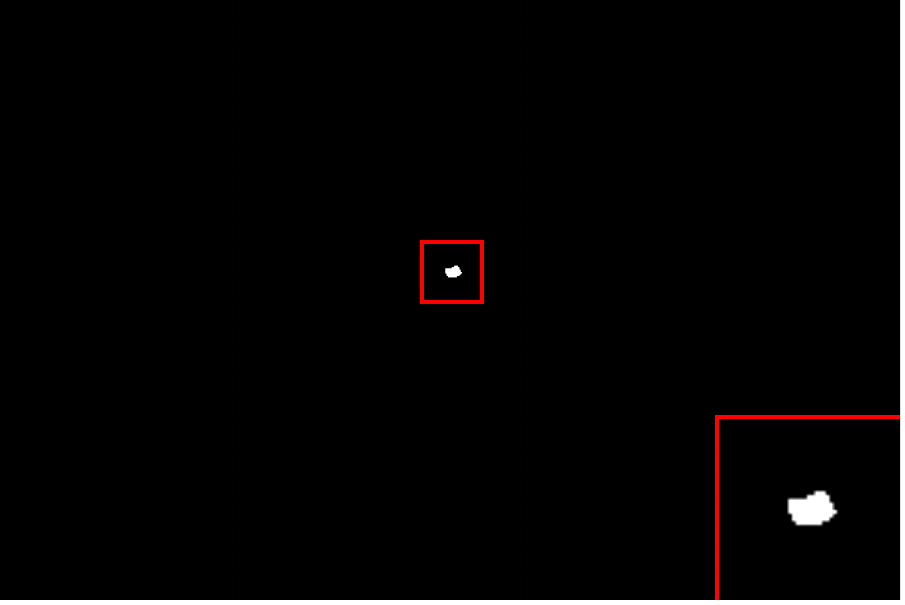}
    \includegraphics[width=\linewidth]{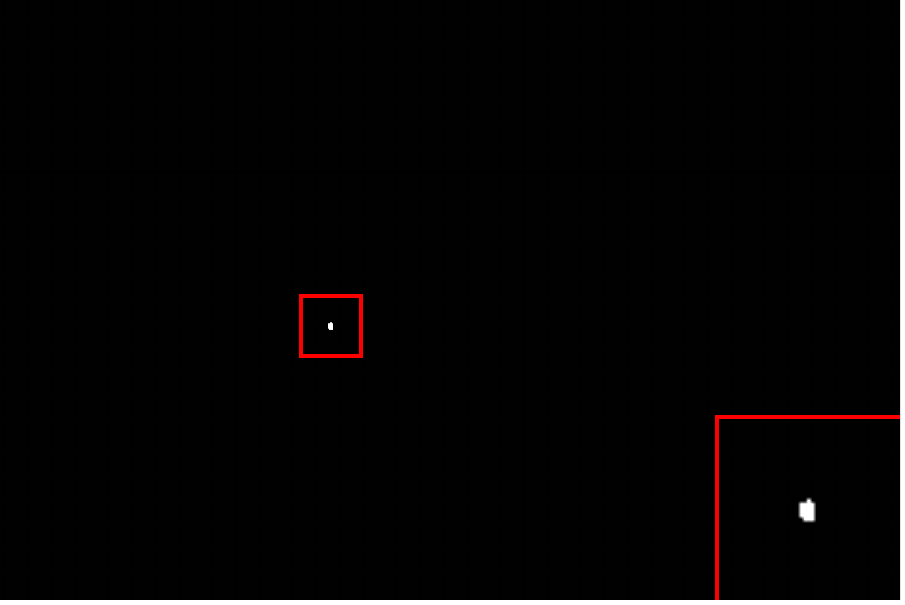}
    \includegraphics[width=\linewidth]{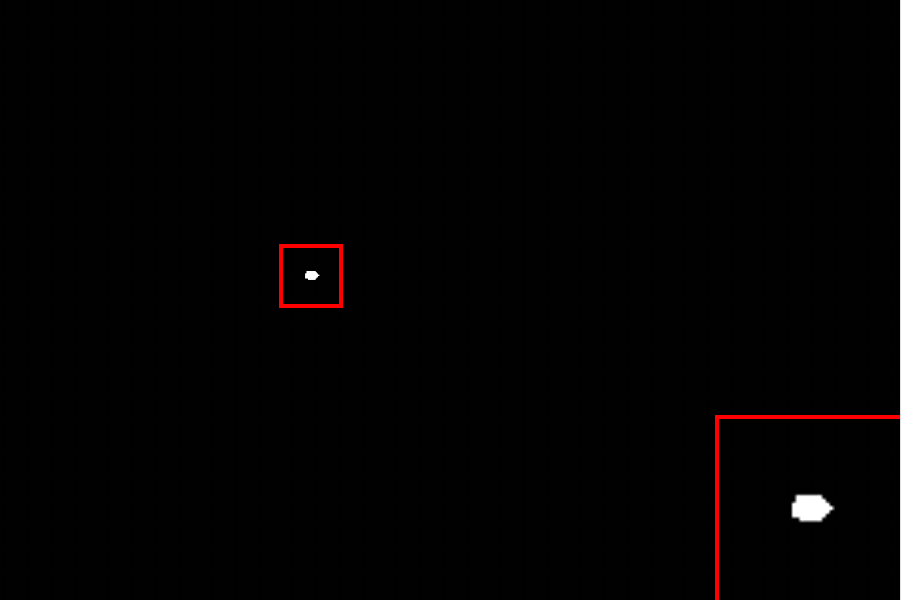}
    \includegraphics[width=\linewidth]{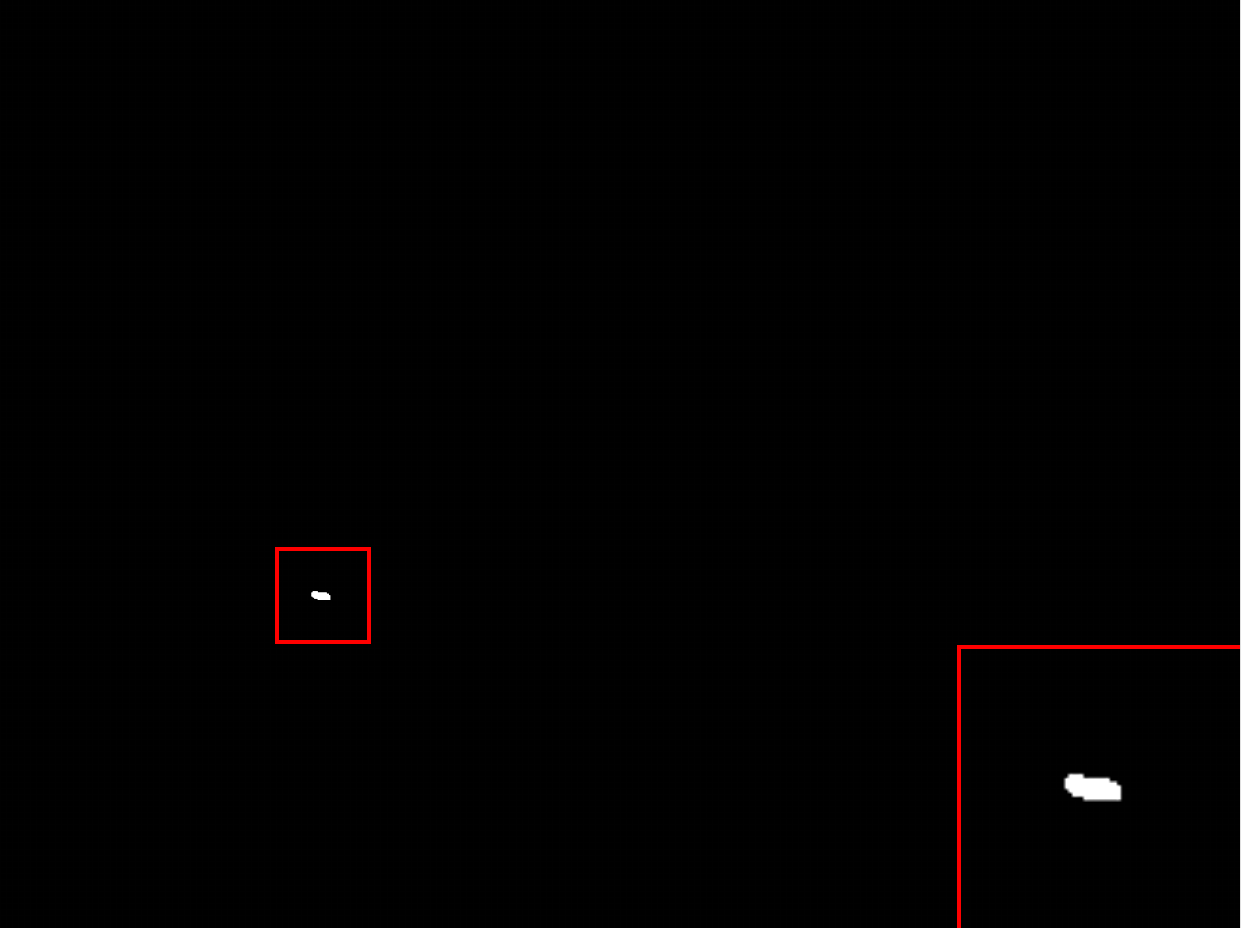}
    \centerline{Ours}
\end{minipage}

\caption{Visual comparison of segmentation results.}
\label{visual_comparison}
\end{figure*}

We compare the proposed method with three types of methods: i) state-of-the-art IRSTS methods, including DNANet \cite{DNA}, UIUNet \cite{UIU}, RDIAN \cite{RDIAN}, and MSHNet \cite{MSH}; ii) in-context learning methods, Painter \cite{Painter} and SegGPT \cite{seggpt}. This type of methods involve concatenating two images along with their masks and applying masked image modeling (MIM) on the mask during training, conditioned on the input images. During inference, models can directly use one pair of images as the input condition to segment other images. iii) training-free generalist one-shot visual reference segmentation models, Matcher \cite{Matcher} and PerSAM \cite{PerSAM}, which belong to the same type as our method. The quantitative results are presented in Table \ref{quantitative_comparison}, showing our method achieves comparable performance
to state-of-the-art IRSTS models trained on large-scale data and significantly outperforms other one-shot segmentation methods. Visual comparison results are shown in Fig. \ref{visual_comparison}.

\subsection{Ablation Study}
\subsubsection{Effect of Proposed Components}
Table \ref{ablation_study} illustrates the effectiveness of each proposed module, PPCF and TLE, by showing performance improvements across $P_d$, $F_a$, and $IoU$ metrics when these modules are incrementally added.

\begin{table}[ht]
\centering
\caption{Effect of proposed components.}
\renewcommand\arraystretch{1}
\begin{tabularx}{\linewidth}{cc YYY}
\hline
TLE & PPCF & $P_d(\uparrow)$ & $F_a(\downarrow)$ & $IoU(\uparrow)$\\
\hline
$\times$ & $\times$ & 90.74 & 63.01 & 25.41\\
$\times$ & $\checkmark$ & 90.74 & 34.87 & 26.22\\
$\checkmark$ & $\times$ & 91.22 & 52.21 & 35.87\\
$\checkmark$ & $\checkmark$ & 95.43 & 31.02 & 49.95\\
\hline
\end{tabularx}
\label{ablation_study}
\end{table}

\subsubsection{Effect of Different Annotations}
Our method allows various types of annotations for the reference image. The results in Table \ref{different_annotation} demonstrate that different types of annotations have a minor influence on the final segmentation results.

\begin{table}[ht]
\centering
\caption{Comparison of different annotations.}
\renewcommand\arraystretch{1}{
\begin{tabularx}{\linewidth}{l YYY}
\hline
Type of Annotation & $P_d(\uparrow)$ & $F_a(\downarrow)$ & $IoU(\uparrow)$\\
\hline
Point & 95.13 & 31.00 & 48.69\\
Bounding Box & 94.80 & 30.79 & 47.42\\
Mask & 95.43 & 31.02 & 49.95\\
\hline
\end{tabularx}
}
\label{different_annotation}
\end{table}

\subsubsection{Effect of Reference Image}
Table \ref{ReferenceImage} demonstrates that our method maintains approximately consistent performance regardless of which frame is chosen as the reference, highlighting the robustness to variations in reference image selection.

\begin{table}[ht]
\centering
\caption{Effect of Reference Image.}
\renewcommand\arraystretch{1}{
\begin{tabularx}{\linewidth}{cYYY}
\hline
\makecell{Reference Image} & $P_d(\uparrow)$ & $F_a(\downarrow)$ & $IoU(\uparrow)$\\
\hline
1st & 95.43 & 31.02 & 49.95\\
10th & 94.43 & 31.57 & 48.90\\
20th & 95.08 & 32.03 & 49.84\\
\hline
\end{tabularx}
}
\label{ReferenceImage}
\end{table}

\subsubsection{Few-Shot Setting}
We also explore the performance of our method in a few-shot setting, as indicated in Table \ref{shotsNumber}. While increasing the number of shots benefits $P_d$, it also impacts $F_a$ and $IoU$ negatively.

\begin{table}[!ht]
\centering
\caption{Effect of the number of reference frames.}
\renewcommand\arraystretch{1}{
\begin{tabularx}{\linewidth}{cYYY}
\hline
Number of shots & $P_d(\uparrow)$ & $F_a(\downarrow)$ & $IoU(\uparrow)$\\
\hline
1 & 95.43 & 31.02 & 49.95\\
2 & 95.61 & 34.90 & 46.36\\
4 & 96.03 & 32.37 & 41.61\\
8 & 96.00 & 38.06 & 43.53\\
\hline
\end{tabularx}
}
\label{shotsNumber}
\end{table}

\subsubsection{Reference-shot from other sequences}
For specific scenes, we may not always have pre-annotated reference images, but this does not mean our method will fail in such cases. Table \ref{otherRef} shows the almost equivalent performance of three sequences using the first frame from other sequences as the reference image. This demonstrates that our method has strong robustness in the selection of reference images.

\begin{table}[ht]
\centering
\caption{Comparison of segmentation performance using first frames from different sequences as reference images.}
\centering
\renewcommand\arraystretch{1.15}{
\begin{tabularx}{\linewidth}{cc YYY}
\midrule
\multicolumn{2}{c|}{\multirow{2}{*}{$P_d$/$F_a$/$IoU$}}& \multicolumn{3}{c}{Target Sequence} \\
\cline{3-5} 
\multicolumn{2}{c|}{}& 1 & 2 & 3\\
\midrule
\multicolumn{1}{c|}{\multirow{3}{*}{\rotatebox{90}{\makecell[c]{Reference\\Sequence}}}} & \multicolumn{1}{c|}{1} & 100/0/36 & 97/3/35 & 100/0/28\\
\multicolumn{1}{c|}{} & \multicolumn{1}{c|}{2} & 100/0/36 & 96/5/61 & 100/0/24\\
\multicolumn{1}{c|}{} & \multicolumn{1}{c|}{3} & 100/6/34 & 96/54/45 & 100/0/23\\
\midrule
\end{tabularx}
}
\label{otherRef}
\end{table}

\section{Conclusion}
In this paper, we propose the first one-shot, training-free method for sequential IRSTS. Our method utilizes SAM, known for its strong generalization, and introduces three modules: LFM, PPCF, and TLE to address the challenges of frame-by-frame manual prompts and domain gaps. Comprehensive experiments demonstrate that our method, requiring only one shot, achieves performance comparable to state-of-the-art methods based on traditional many-shot supervision and significantly outperforms other one-shot segmentation methods.

\end{document}